\pdfoutput=1

\documentclass[11pt]{article}

\usepackage{EMNLP2023}
\usepackage{svg}
\usepackage{amsmath}
\usepackage{amssymb}
\usepackage{multicol,multirow}
\usepackage{xcolor}
\usepackage{booktabs}
\usepackage{caption}
\usepackage{hyperref}

\usepackage{times}
\usepackage{latexsym}

\usepackage[T1]{fontenc}

\usepackage[utf8]{inputenc}

\usepackage{microtype}

\usepackage{inconsolata}

\title{Large Language Models as Neurolinguistic Subjects: \\Discrepancy between Performance and Competence}

\author{Linyang He\textsuperscript{1, 2}\qquad Ercong Nie\textsuperscript{3, 4}\qquad Helmut Schmid\textsuperscript{4}\\ 
\textbf{~Hinrich Sch\"utze\textsuperscript{3, 4}\qquad Nima Mesgarani\textsuperscript{1}\qquad Jonathan  Brennan\textsuperscript{2}}\\
\textsuperscript{1}Columbia University~\textsuperscript{2}University of Michigan\\
\textsuperscript{3}Munich Center for Machine Learning, Germany ~\textsuperscript{4}LMU Munich, Germany\\
\texttt{linyang.he@columbia.edu} \qquad 
\texttt{\{nie,schmid\}@cis.lmu.de}\\
\texttt{hinrich@hotmail.com}\qquad \texttt{nima@ee.columbia.edu}\qquad \texttt{jobrenn@umich.edu}
}

\begin{document}
\maketitle

\begin{abstract}
This study investigates the linguistic understanding of Large Language Models (LLMs) regarding signifier (form) and signified (meaning) by distinguishing two LLM assessment paradigms: psycholinguistic and neurolinguistic. Traditional psycholinguistic evaluations often reflect statistical rules that may not accurately represent LLMs' true linguistic competence. We introduce a neurolinguistic approach, utilizing a novel method that combines minimal pairs and diagnostic probing to analyze activation patterns across model layers. This method allows for a detailed examination of how LLMs represent form and meaning, and whether these representations are consistent across languages. We found: (1) Psycholinguistic and neurolinguistic methods reveal that language performance and competence are distinct; (2) Direct probability measurement may not accurately assess linguistic competence; (3) Instruction tuning won't change much competence but improve performance; (4) LLMs exhibit higher competence and performance in form compared to meaning. Additionally, we introduce new conceptual minimal pair datasets for Chinese (COMPS-ZH) and German (COMPS-DE), complementing existing English datasets.\footnote{Code and data available \href{https://github.com/LinyangHe/LLM-Neurolinguistic-Subject}{here}.}
\end{abstract}

\section{Introduction}
Large Language Models (LLMs) have demonstrated remarkable reasoning, linguistic, arithmetic, and other cognitive abilities. The advent of LLMs has reignited cross-disciplinary discussions about what sorts of behavior are ``intelligence'', even if the intelligence exhibited by LLMs may differ from human intelligence \cite{sejnowski2023large}.  LLMs have drawn the attention of researchers from various fields, including linguistics, cognitive science, computer science, and neuroscience, who investigate how LLMs develop and exhibit these capabilities.

\begin{figure}[t]
    \centering
    \includegraphics[width=0.9\linewidth]{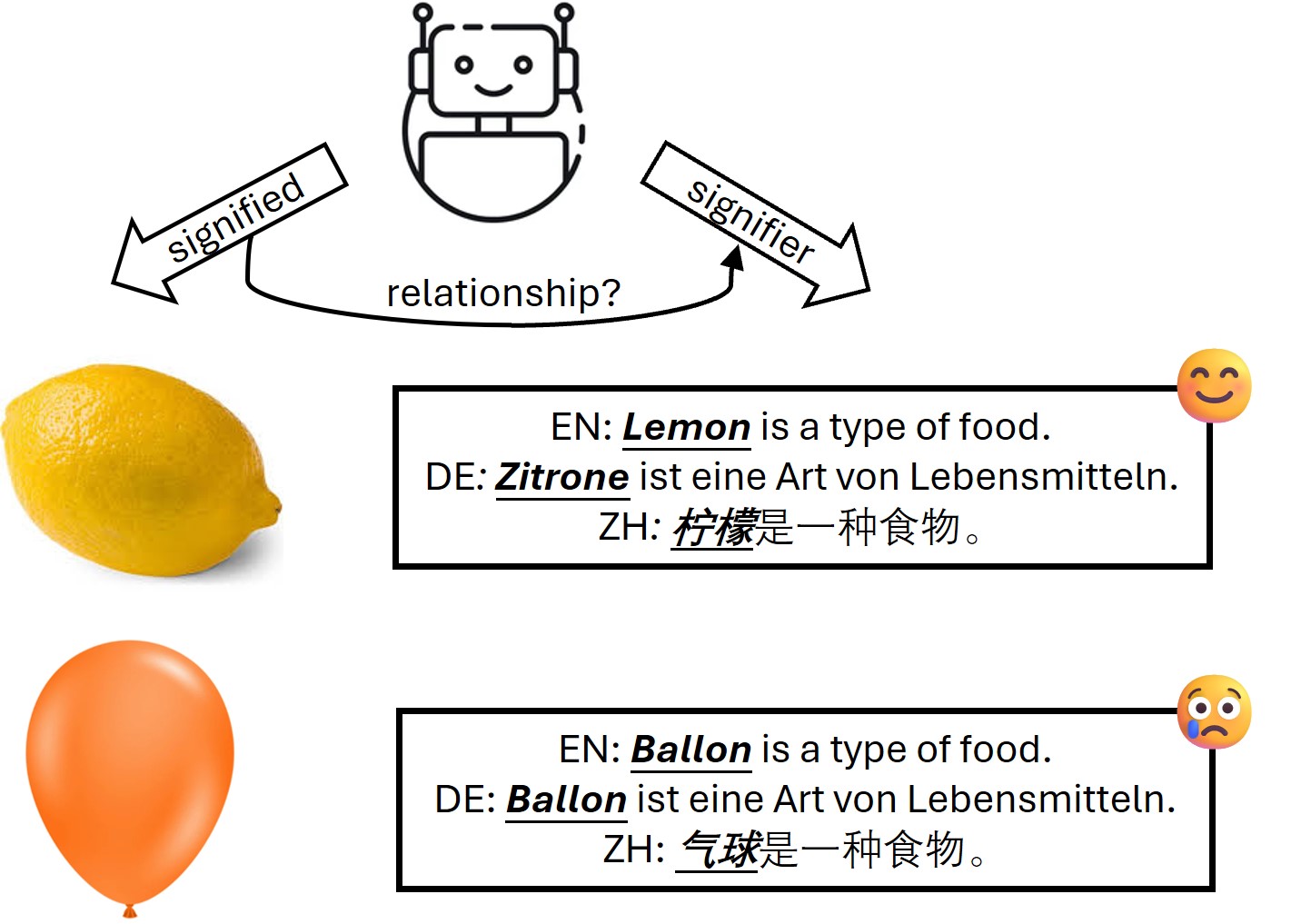}
    \caption{Illustration of LLMs processing the same signified (meaning) across different signifiers (forms). 
    }
    \label{fig:research_question}
\end{figure}

\begin{figure*}[t]
    \centering
    \includegraphics[width=0.9\linewidth]{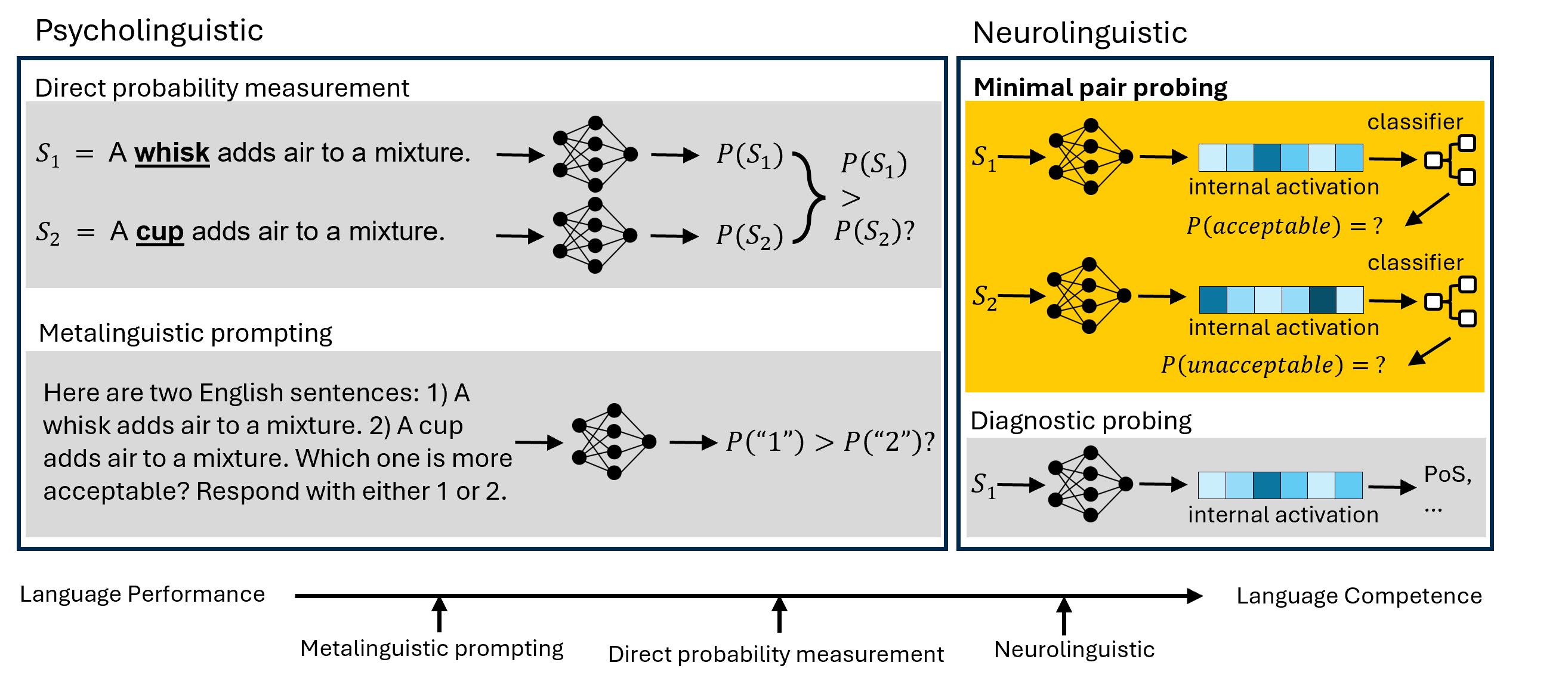}
    \caption{Psycholinguistic vs. Neurolinguistic Paradigm. Both direct probability measurement and metalinguistic prompting can be considered as psycholinguistic methods, while minimal pair probing \cite{he2024decoding} and other diagnostic probing are neurolinguistic.}
    \label{fig:framework}
\end{figure*}


There is currently a heated debate about whether LLMs understand human language or whether their performance is simply the product of complex statistical relationships~\cite{mitchell2023debate}. 
A central aspect of this debate concerns the nature of LLMs' linguistic representations.
Using the semiotic framework of language proposed by~\citet{de1989cours}, which distinguishes between the signifier (form) and the signified (meaning), we can inquire into the extent to which LLMs comprehend the form and meaning, and how form and meaning intertwist with each other. 
Is LLMs‘ understanding of language meaning merely a statistical outcome based on their grasp of language form? 
When different languages express a shared concept with distinct forms, do LLMs create similar representations for these variations? 
How can we better understand the representations of form and meaning in these systems that support the observed patterns of performance? 

The underlying processes remain unclear due to the opaque nature of neural networks. Therefore, we need appropriate methods to assess their true linguistic understanding. 

Drawing inspirations from the cognitive study on human language processing, we propose that the assessment of LLMs can be divided into two primary paradigms: \textit{psycholinguistic} and \textit{neurolinguistic}. 
As illustrated in Figure \ref{fig:framework}, the
psycholinguistic paradigm measures the model's output probabilities, directly reflecting the model's behavior and performance. The neurolinguistic paradigm delves into the internal representations of LLMs.



When treating LLMs as psycholinguistic subjects, their responses may leverage their grasp of form, relying on statistical correlations, to create an illusion of understanding meaning. This enables LLMs to produce structurally coherent but not necessarily semantically accurate responses, as their ``understanding'' is shaped by patterns rather than true conceptual processing \cite{harnad1990symbol, bender2020climbing, nie2024decomposed}. Consequently, psycholinguistic evaluations tend to reflect performance rather than competence, as they assess external outputs that may not fully capture the underlying linguistic knowledge encoded within the model. This mismatch suggests that psycholinguistic evaluation results might not accurately represent the true linguistic competence of LLMs.

In contrast, examining LLMs as neurolinguistic subjects focuses on internal representations, providing a more direct assessment of competence by moving beyond surface-level biases \cite{firestone2020performance}. To achieve this, we adapted the decoding probing method by \citet{he2024decoding}, referred to as ``minimal pair probing'', to analyze how LLMs encode form and meaning across layers. This approach allows for a finer distinction between performance and competence, revealing insights that psycholinguistic methods might overlook.

In order to address questions about whether LLMs maintain consistent underlying representations of the same concept when the form changes across multiple languages, we also create a multilingual minimal pair dataset (COMPS-ZH for Chinese and COMPS-DE for German).

By evaluating LLMs in both psycholinguistic and neurolinguistic paradigms, we found: 
1) Psycholinguistic and neurolinguistic results reveal very different patterns, suggesting both paradigms are necessary for a comprehensive understanding of LLMs.
2) Though more intrinsic than metalinguistic prompting, direct probability measurement may still not accurately assess linguistic competence, as it remains influenced by statistical patterns.
3) LLMs acquire competence in linguistic form more easily, earlier, and with greater accuracy than in meaning.
4) As linguistic form varies across languages, LLMs' understanding of the same concept shifts accordingly, with meaning competence linearly correlated to form. This suggests that signifier and signified in LLMs may not be independent, and maintaining conceptual representations likely depends on statistical correlations with form.


 
\section{Psycholinguistic vs. Neurolinguistic Paradigm}
\subsection{Cognitive Science Background}
Psycholinguistics and neurolinguistics offer distinct yet complementary perspectives on human language processing. Psycholinguistics focuses on the psychological and cognitive processes that enable humans to understand and use language \cite{field2004psycholinguistics, traxler2011handbook}. In contrast, neurolinguistics explores the underlying neural mechanisms and brain structures involved in language processing \cite{friederici2011brain, brennan2022language, kemmerer2022cognitive}. Both paradigms offer a valuable model for probing the linguistic capacities and potential intelligence of LLMs. 

\subsection{In LLM Assessment Research}
\subsubsection*{Psycholinguistic paradigm: direct probability measurement and metalinguistic prompting}
Recent studies often use prompting to evaluate the linguistic capabilities of LLMs. 
These implicit tests were referred to as \textit{metalinguistic judgments} by \citet{hu2023prompting}. 
However, it is important to note that the performance of LLMs in specific linguistic prompting tasks only indirectly reflects their internal linguistic representations due to the inherent limitations of such prompting tasks: an LLM chat system might give a ``reasonable'' response just because of the statistical relationships between prompt and reply \cite{hofstadter1995fluid}. \citet{hu2023prompting} argue that
it is uncertain whether the LLMs' responses to metalinguistic prompting align with the underlying internal 
representations.

Computing a model's probability of generating two minimally
different sentences is one way to address these concerns \cite{hu2023prompting}.
The minimal difference between the two sentences (e.g., replacement of a single word) makes one sentence acceptable while the other is not \cite{linzen2016assessing}. Here are two examples for testing grammatical and conceptual understanding, respectively:\\
(1) \textit{Simple agreement} \cite{warstadt2020blimp}: 
\begin{tabbing}
\hspace*{0.5cm}a. \=  The cats \underline{annoy} Tim. (\textit{acceptable})\\
\hspace*{0.5cm}b. \> *The cats \underline{annoys} Tim. (\textit{unacceptable})
\end{tabbing}
(2) \textit{Concept understanding} \cite{misra2023comps}: 
\begin{tabbing}
\hspace*{0.5cm}a. \=  A \underline{whisk} adds air to a mixture. (\textit{acceptable})\\
\hspace*{0.5cm}b. \> *A \underline{cup} adds air to a mixture. (\textit{unacceptable})
\end{tabbing}

A language model is considered to perform correctly on this task if it assigns a higher probability to the acceptable sentence compared to the unacceptable one \cite{marvin-linzen-2018-targeted}. Researchers have created syntactic, semantic/conceptual, and discourse inference tasks for the minimal pair method. They provide more precise insights into the abilities of LLMs compared to metalinguistic prompting \cite{futrell2019neural, gauthier2020syntaxgym, hu2020systematic, warstadt2020blimp, beyer2021incoherence, misra2023comps, kauf2023event}.

Through either metalinguistic judgement or direct probability measurement methods, these tasks essentially treat LLMs as \textit{psycholinguistic} subjects \cite{futrell2019neural}. This research paradigm resembles cognitive psychology by having LLMs perform tasks, such as cloze and question answering, and then evaluating their performance without examining the internal representations, 
in a manner similar to how subjects participate in psychological experiments.
Information about the inner workings of a model is inferred either from its output or from the probabilities it assigns to different possible outputs. The internal states of the LLM (i.e.\ its intermediate layers) are not examined.


\subsubsection*{Neurolinguistic paradigm: diagnostic probing}
Another line of research focuses on studying the internal representations, emphasizing a \textit{neurolinguistic} approach to understanding LLMs. Essentially, diagnostic probing methods in evaluating language models can be considered as neurolinguistic paradigms as they examine the internal states of LMs \cite{belinkov2019analysis, belinkov2022probing}, while the term `\textit{neurolinguistic}' hasn't been applied to the field before. Diagnostic probing involves training a classifier to predict linguistic properties from the hidden states of LMs. Following this paradigm, researchers decode syntactic, semantic, morphological, and other linguistic properties from the hidden states of LMs \cite{kohn2015s, gupta2015distributional, shi2016does, tenney2019you, hewitt2019structural, manning2020emergent}. 

\section{Minimal Pair Probing = Minimal Pair + Diagnostic Probing}
\label{sec:minimal}
\begin{table*}[th]
    \centering
    \scalebox{0.8}{
    \begin{tabular}{lllll}
    \hline
    \textbf{Minimal Pair} & \textbf{Duality} & \textbf{Language} & \textbf{\# of Pair} & \textbf{Description}                     \\ \hline
    BLiMP                  & Form          & English           & 67, 000             & 67 tasks across 12 grammatical phenomena \\
    CLiMP                  & Form          & Chinese           & 16, 000             & 16 tasks across 9 grammatical phenomena  \\
    DistilLingEval         & Form          & German            & 8, 000              & 8 German grammatical phenomena           \\
    COMPS                  & Meaning       & English           & 49, 340             & 4 types of conceptual relationship       \\
    COMPS-ZH               & Meaning       & Chinese           & 49, 340             & 4 types of conceptual relationship       \\
    COMPS-DE               & Meaning       & German            & 49, 340             & 4 types of conceptual relationship       \\ \hline
    \end{tabular}}
    \caption{Overview of datasets in our study.}
    \label{tab:dataset}
\end{table*}

\begin{table*}[h]
    \centering
    \scalebox{0.8}{
    \begin{tabular}{lll}
    \midrule
    Duality & Method & Example \\
    \hline
      \multirow{3}{*}{Form}   & Direct & \{\textcolor{blue}{Mice are hurting a waiter}, \textcolor{red}{Mice was hurting a waiter}\} \\
      & \multirow{2}{*}{Meta} & Here are two English sentences: 1) Mice are hurting a waiter. 2) Mice was hurting a waiter. Which \\
      & & sentence is a better English sentence? Respond with either 1 or 2 as your answer. Answer: \{\textcolor{blue}{1}, \textcolor{red}{2}\} \\
      \multirow{3}{*}{Meaning}   & Direct & \{\textcolor{blue}{Helmet can absorb shocks}, \textcolor{red}{Cap can absorb shocks}\} \\
      & \multirow{2}{*}{Meta} & What word is most likely to come next in the following sentence (helmet, or cap)? What can absorb \\
      & &  shocks? \{\textcolor{blue}{helmet}, \textcolor{red}{cap}\}\\
    \bottomrule
    \end{tabular}}
    \caption{Prompt examples for baseline methods. The region where we measure probability is marked in color. Correct sentences and answers are in \textcolor{blue}{blue}; incorrect in \textcolor{red}{red}.}
    \label{tab:prompts}
\end{table*}

While prior neurolinguistic approaches have explored internal representations, they often employed coarse-grained datasets and primarily focused on decoding linguistic labels from embeddings, providing a general perspective on the linguistic features encoded in LMs. In contrast, the minimal pair probing method presented by \citet{he2024decoding} integrates minimal pair design with diagnostic probing. This combination leverages the granularity of minimal pair design and the layer-wise insights of diagnostic probing, thereby enabling a more detailed analysis of internal patterns for form and meaning. We adopt minimal pair decoding as the neurolinguistic paradigm in our work.

Specifically, given an LLM $f:x_{0, 1, ..., i} \rightarrow x_{i+1}$ trained on dataset $\mathcal{D}_O$, we can extract the hidden state representations $f_l(S)$ of the $l$-th layer of stimuli $S$. Given a minimal pair dataset $\mathcal{D}_m$ = $\{(S_+^{i}, S_-^{i}), (z_+^{i},z_-^{i})\}$ with each sentence $S$ has a label $z$, we have internal representation $f_l(S_+^{i})$ and $f_l(S_-^{i})$ for each sentence. A minimal probing classifier $g: f_l(S) \rightarrow \hat{z}$ is trained and evaluated on $\mathcal{D}_m$, with grammatical/conceptual performance measure $\mathrm{Perf}(f, \mathcal{D}_O, g, \mathcal{D}_m)$. 


Note that our focus is on evaluating the linguistic competence of the LLM $f$ itself, i.e., $\mathrm{Perf}(f, \mathcal{D}_O)$, rather than the capacity of the probing classifier $g$. As suggested by \citet{hewitt2019designing}, even untrained or random representations can yield surprisingly high probing accuracy, raising concerns that the classifier may exploit dataset artifacts rather than meaningful representations. To control for the potential bias introduced by $g$, we construct a random embedding baseline. 

Specifically, for each sentence in the dataset, we assign a fixed random vector $r$, sampled from a Gaussian distribution with the same mean and standard deviation as the real model embeddings $f_l(S)$. Importantly, each sentence is consistently assigned the same random vector across occurrences, preserving instance-level identity that the probing classifier might exploit. This allows us to assess the extent to which task performance can be driven by superficial sentence-level cues rather than meaningful representations. We then compute $\mathrm{Perf}(g, \mathcal{D}_m)$ by training $g$ on these random embeddings, which reflects the inherent predictability or “shortcut” potential of the probing task. Therefore, our performance score incorporates a correction factor based on this random baseline, defined as:

\begin{equation}
\small
\mathrm{Perf}(f, \mathcal{D}_O) \triangleq \mathrm{Perf}(f, \mathcal{D}_O, g, \mathcal{D}_m) \cdot (1 +\frac{ 0.5- \mathrm{Perf}(g, \mathcal{D}_m)}{0.5})
\label{eq:perf}
\end{equation}

This formula applies a correction term, penalizing cases where the probing classifier performs well even on random embeddings. When $\mathrm{Perf}(g, \mathcal{D}_m) = 0.5$, the correction factor is 1; if the performance is higher, the factor shrinks toward 0, discouraging overfitting or trivial tasks; if it drops below 0.5, the factor exceeds 1, slightly amplifying the model’s score. This ensures that only meaningful representations in $f$ contribute to the final evaluation.

\section{Experiment Setup}


\subsection{Datasets and Models}
We use minimal pair probing for English, Chinese, and German to assess grammaticality (form) and conceptuality (meaning). Table \ref{tab:dataset} presents the overall dataset information used in our experiments. We use Llama2-7B, Llama3-8B, and Qwen-7B models in both base and chat versions. Further dataset and model descriptions are in Appendix \ref{sec:appendix_dataset} and \ref{sec:appendix_models}.

\begin{figure*}[t]
    \centering
    \includegraphics[width=1.0\linewidth]{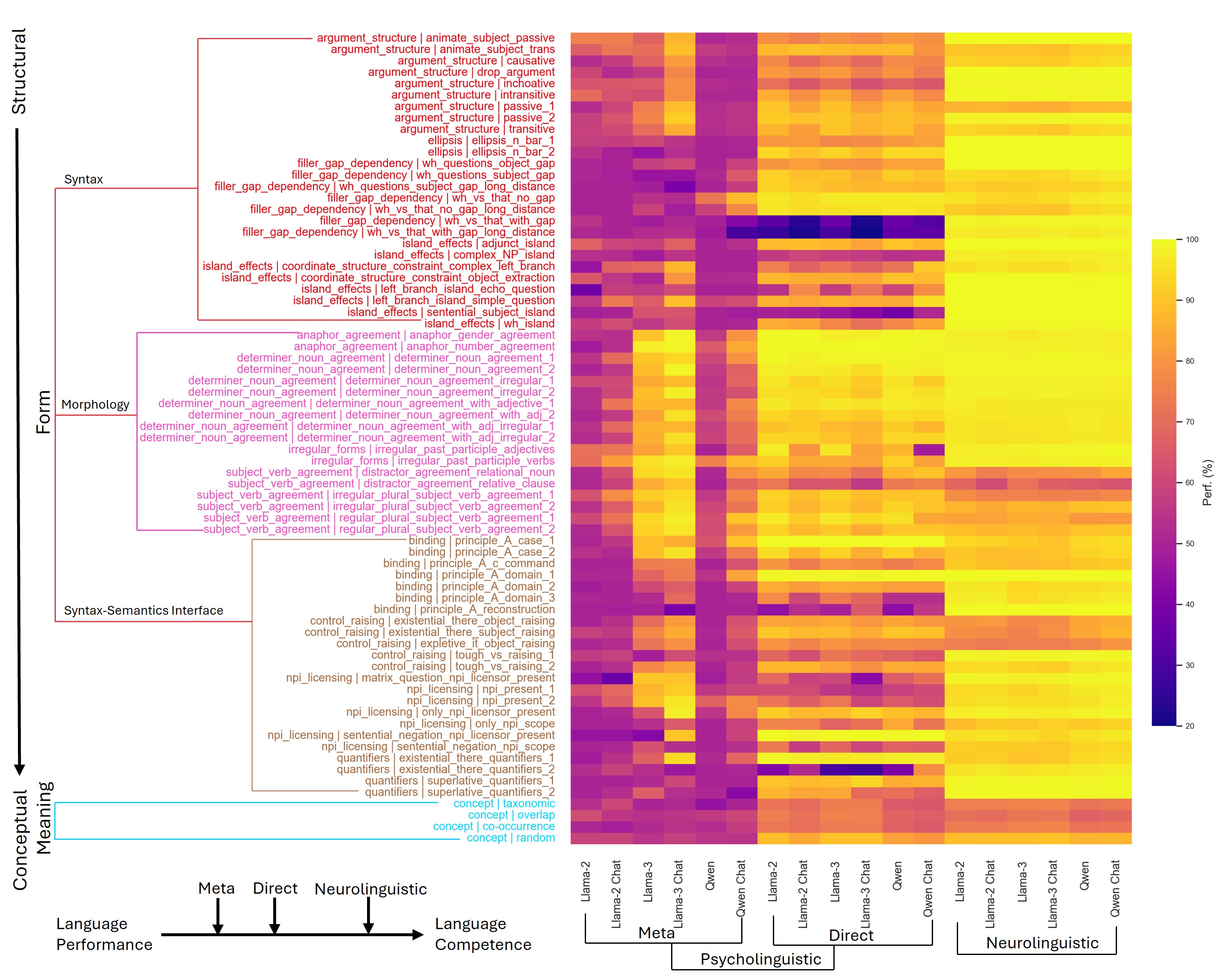}
    \caption{Psycholinguistic (meta and direct) and neurolinguistic performance across models and linguistic tasks. The x-axis represents different models and conditions (base and chat), while the y-axis categorizes linguistic tasks based on structural (syntax, morphology, syntax-semantics interface) and conceptual (meaning) levels.}
    \label{fig:grand_results}
\end{figure*}

\subsection{Setup for Psycholinguistic Analysis}
\paragraph{Direct} 
Direct probability measurement calculates the probability of a sentence based on model logits. Accuracy is determined by whether the model assigns a higher probability to the grammatically or conceptually correct sentence within the minimal pair.

\paragraph{Meta} 
Metalinguistic prompting involves explicitly asking a question or specifying a task that requires a judgement about a linguistic expression. Following \citet{hu2023prompting}, we use one prompt for a minimal pair to present both sentences at once. For form tasks, we assign an identifier (1 or 2) to each sentence in the pair, present a multiple-choice question comparing both sentences, and compare the probabilities assigned by the model to each answer option, ``1'' or ``2''. 
For meaning tasks, we reformulate the property into a question and compare the probabilities of acceptable and unacceptable concepts as sentence continuations. Table \ref{tab:prompts} presents the prompts used in the experiments.


\subsection{Setup for Neurolinguistic Analysis}
\paragraph{Sentence Embedding} We extract the last token in each sentence from each layer to serve as the representation for the whole sentence. Last token pooling ensures the representation contains the information of all preceding tokens \cite{SFRAIResearch2024}.
\paragraph{Probing Performance} We use logistic regression as the probing classifier and F1 score as the evaluation metric. The score for $\mathrm{Perf}(f, \mathcal{D}_O, g, \mathcal{D}_m)$ and $\mathrm{Perf}(g, \mathcal{D}_m)$ is calculated as the average F1 score across 5 cross-validation folds. Final performance $\mathrm{Perf}(f, \mathcal{D}_O)$ is given by Formula \ref{eq:perf}.
\paragraph{Saturation and Maximum Layer} We define the feature learning Saturation Layer as the layer where performance first reaches 95\% of the peak on the curve. This layer indicates the number of layers required for the model to adequately learn specific linguistic features, after which its ability to capture these features stabilizes. The Maximum Layer is the layer at which performance reaches its peak.
\paragraph{Unsupervised Analysis} We use t-SNE to visualize the sentence embedding of Llama2-7B for English form tasks. We employ PCA to reduce the dimensionality of the sentence embedding to 50 before applying t-SNE.

\section{Results}

\subsection{Psycholinguistic vs Neurolinguistic}

\begin{figure}[h]
    \centering
    \includegraphics[width=0.9\linewidth]{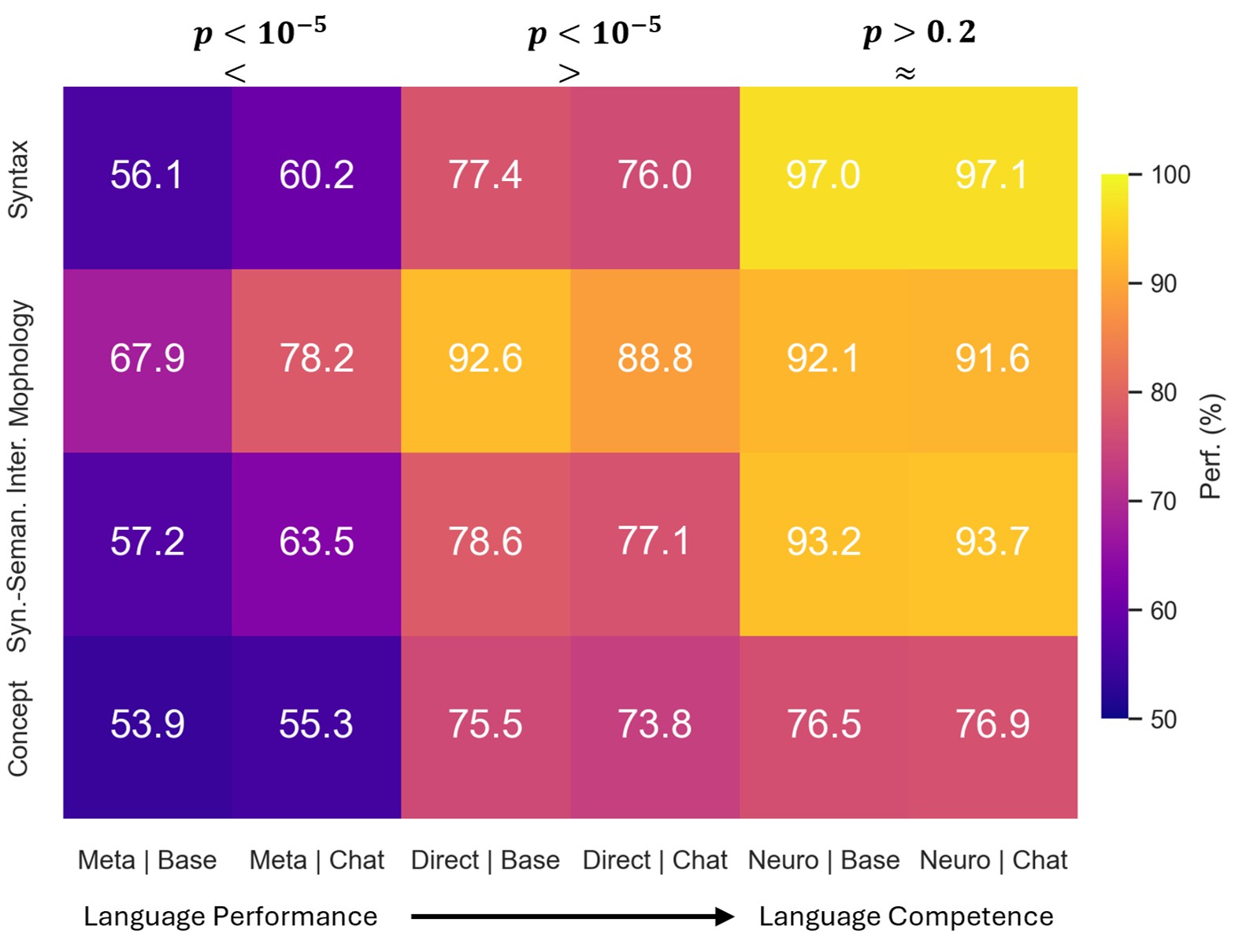}
    \caption{
    Averaged psycholinguistic (meta and direct) and neurolinguistic results across models and tasks. t-tests were conducted on the original (pre-averaging) results between base and chat models, with p-values annotated. 
    }
    \label{fig:form_meaning_en_a}
\end{figure}

\begin{figure}[h]
    \centering
    \includegraphics[width=0.9\linewidth]{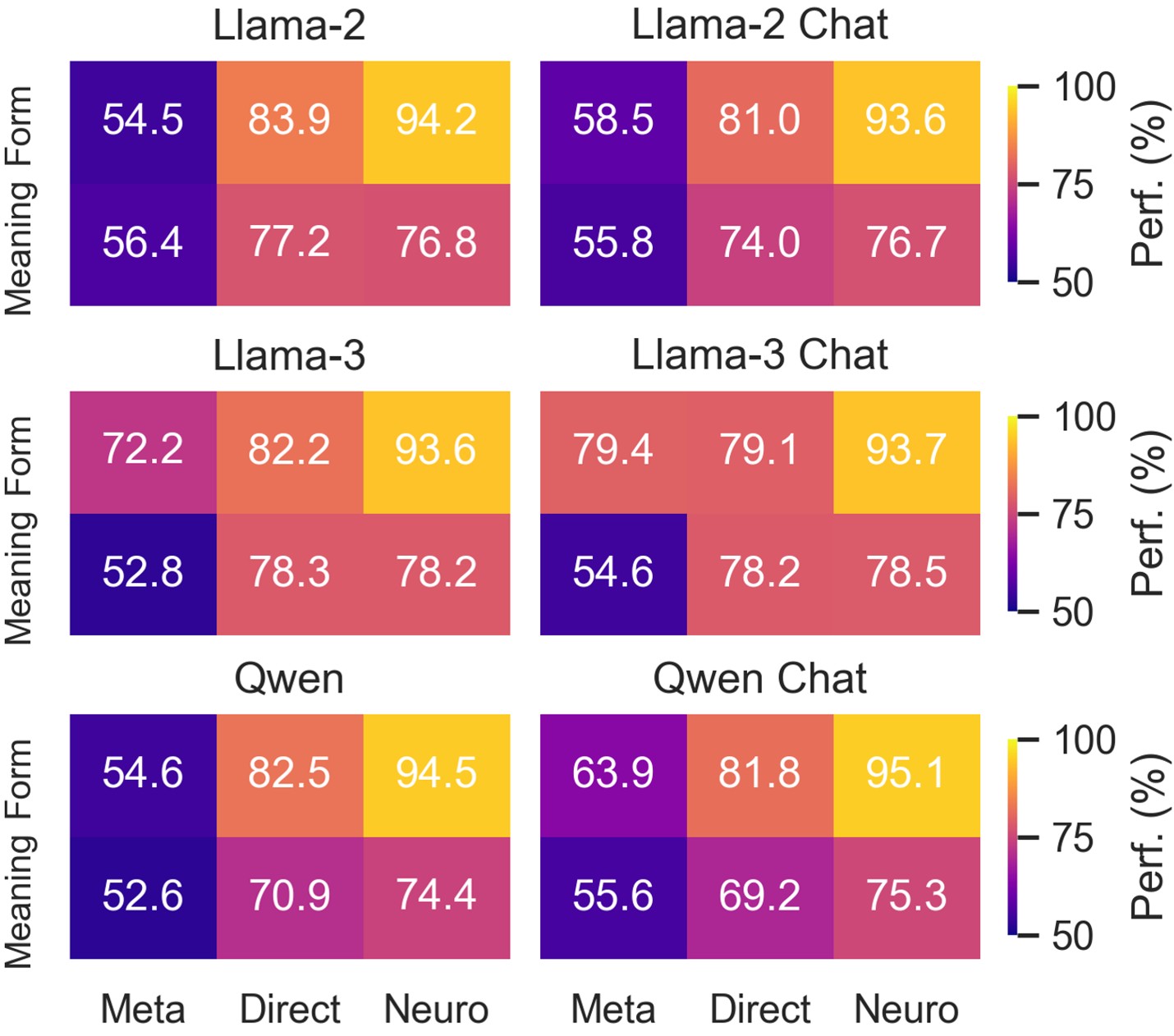}
    \caption{
    Psycholinguistic and neurolinguistic performance for form (morphology, syntax-semantics interface, and syntax) and meaning (concept). 
    }
    \label{fig:form_meaning_en_b}
\end{figure}

Figure \ref{fig:grand_results} shows the performance of LLMs across all linguistic tasks. Figure \ref{fig:form_meaning_en_a} demonstrates the averaged performance of LLMs across models and 4 levels (syntax, morphology, syntax-semantics interfaces, concept). Figure \ref{fig:form_meaning_en_b} presents the average performance of LLMs across form and meaning tasks for Direct, Meta, and Neuro\footnote{We refer to minimal pair probing as Neuro for simplicity.} methods.
We use the last layer's performance in the Neuro method when comparing psycho- and neurolinguistic paradigms, as both direct probability measurement and metalinguistic prompting rely on the last layer of LLMs.

\paragraph{Language performance and competence are distinct (Competence $>$ Performance).} Figure \ref{fig:form_meaning_en_a} and \ref{fig:form_meaning_en_b} shows distinct results between language performance and competence. Moving from Meta → Direct → Neuro, the evaluation focus gradually shifts from language performance (task execution ability) to language competence (the underlying linguistic ability). Within the same task category, Neuro methods consistently yield higher performance than Direct methods, which in turn outperform Meta methods. This indicates that when evaluating pure linguistic competence, LLMs perform well, but their performance drops when assessed in a task-based setting.

\paragraph{Tasks that emphasize language performance become more difficult, even if their language competence is high.} For example, in the Neuro setting, performance on Syntax tasks reaches 97\%, while in the Meta setting, it drops to 56.1\%, showing a significant gap. This suggests that even when an LLM has strong competence in a given task, its performance can significantly decline when assessed under a performance-oriented evaluation.

\paragraph{Direct probability measurement might not be a true competence assessment.} As the Neuro method measures the internal representations of LLMs directly, it could serve as a reliable ground truth for estimating linguistic competence. Direct probability measurement falls short of achieving this ground truth in form assessment (especially for syntax and syntax-semantics-interface as shown in Figure \ref{fig:form_meaning_en_b}). 

\begin{figure}[h]
    \centering
    \includegraphics[width=0.8\linewidth]{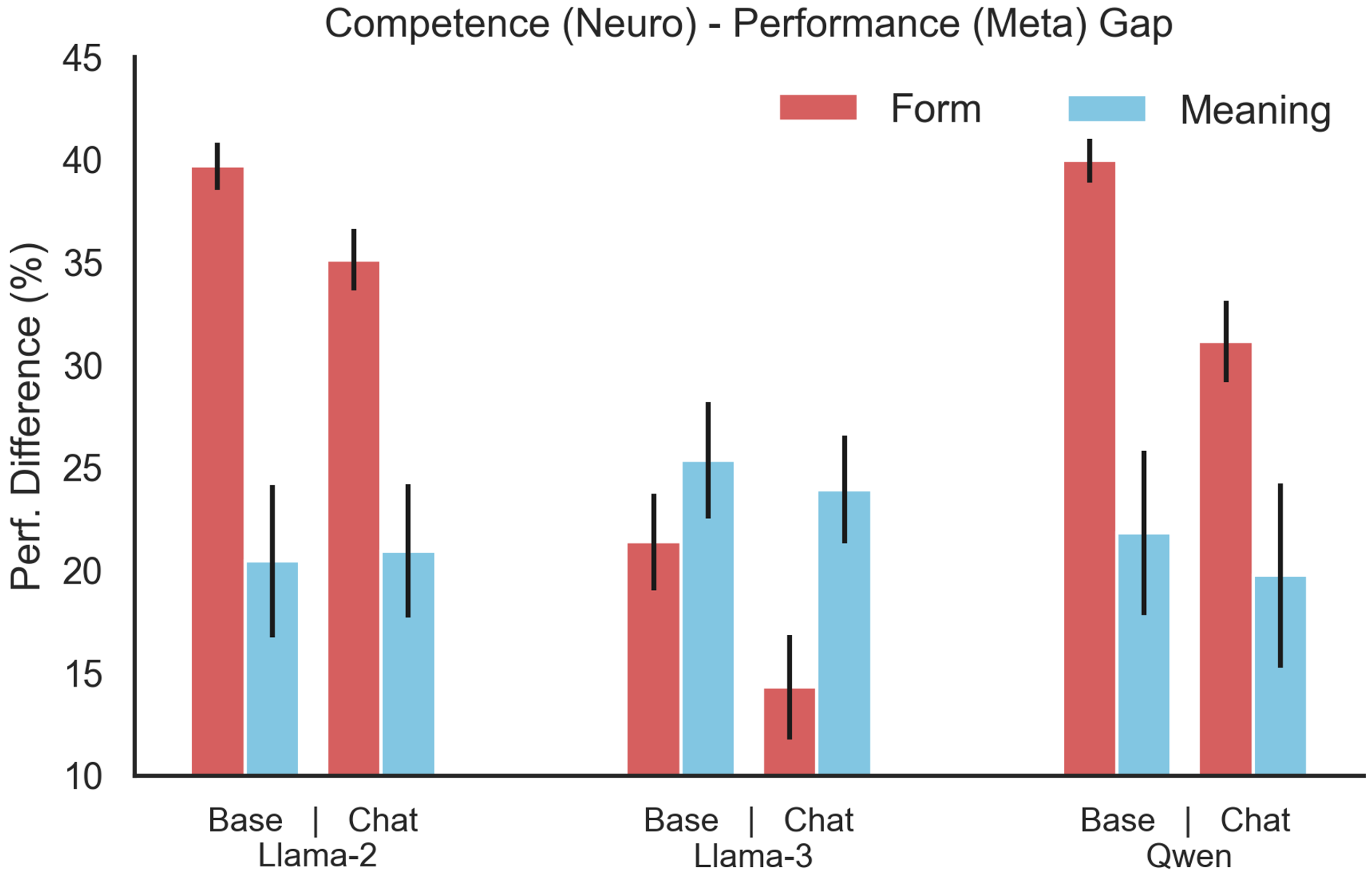}
    \caption{Competence and performance gap drops after instruction tuning.}
    \label{fig:gap_bar}
\end{figure}

\begin{figure*}[t]
    \centering
    \includegraphics[width=1.0\linewidth]{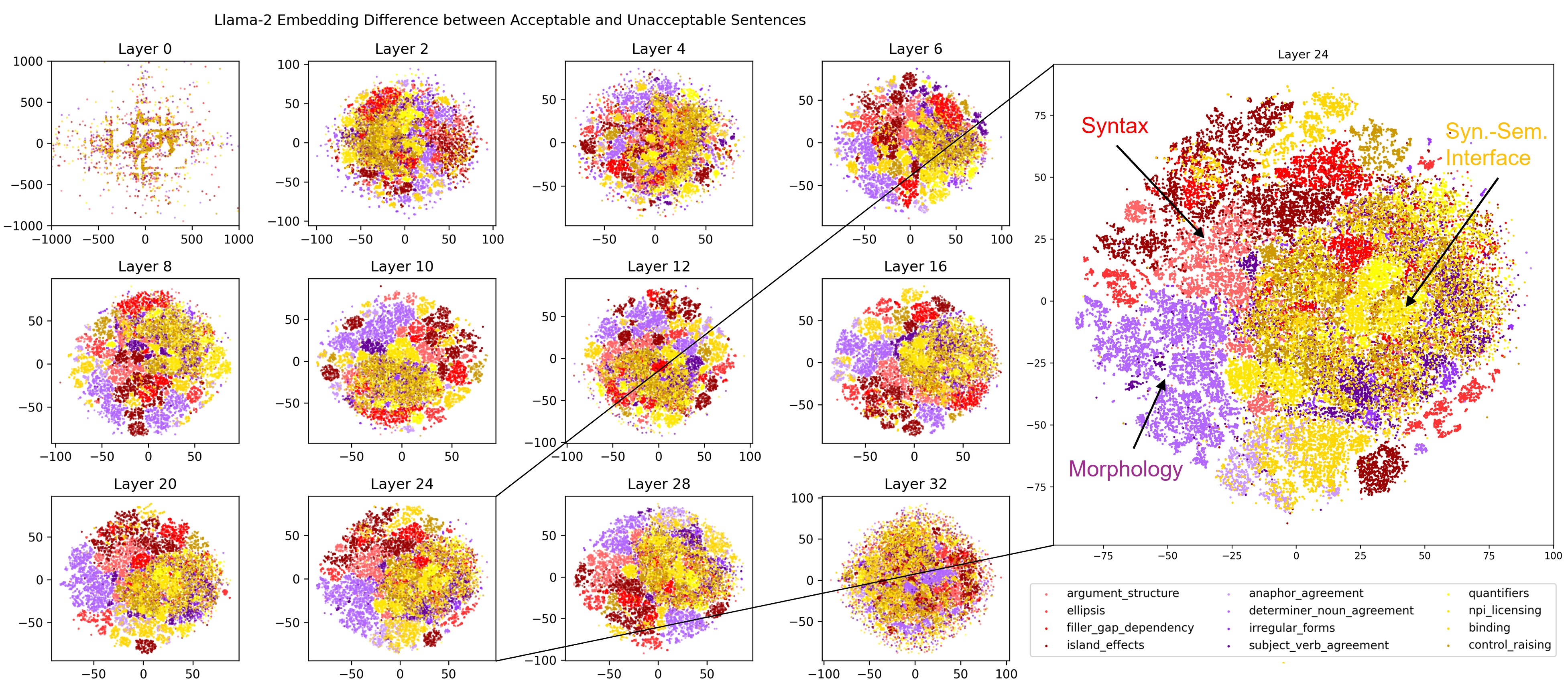}
    \caption{t-SNE visualization of embedding differences between acceptable and unacceptable sentences, with red for syntax, purple for morphology, and yellow for the syntax-semantics interface.}
    \label{fig:tsne}
\end{figure*}

\paragraph{LLMs exhibit stronger mastery of form than meaning, regardless of performance or competence.} As shown in Figure \ref{fig:form_meaning_en_b}, LLMs consistently perform better on form-related tasks than on meaning-related tasks. This trend holds regardless of whether the model is a base or chat version. Crucially, this pattern is evident across all evaluation methods. This indicates that LLMs have a stronger grasp of linguistic form than conceptual meaning, whether assessed through task execution or underlying capability.

\paragraph{Instruction tuning won't change much competence but improve performance.} Neuro results between the base and chat versions of LLMs reveal that instruction fine-tuning does not significantly alter the language competence of the models (t-test between Neuro-Base vs. Neuro-Chat as shown in Figure \ref{fig:form_meaning_en_a}). With instruction fine-tuning (chat versions of LLMs), Meta performance on form improves significantly while meaning understanding remains stable. Figure \ref{fig:gap_bar} illustrates that after instruction tuning, the competence-performance gap (Neuro-Meta) significantly decreases for form-related tasks, while the change for meaning-related tasks remains relatively small. This indicates that fine-tuning with well-designed instructions helps LLMs improve their performance on form-related tasks, bringing them closer to their underlying competence. However, for meaning-related tasks, instruction tuning does not lead to a fundamental improvement in understanding. This indicates that more optimized information access strategies can enhance the external performance of language models, particularly for form-related tasks.

\begin{figure*}[h]
    \centering
    \includegraphics[width=1.0\linewidth]{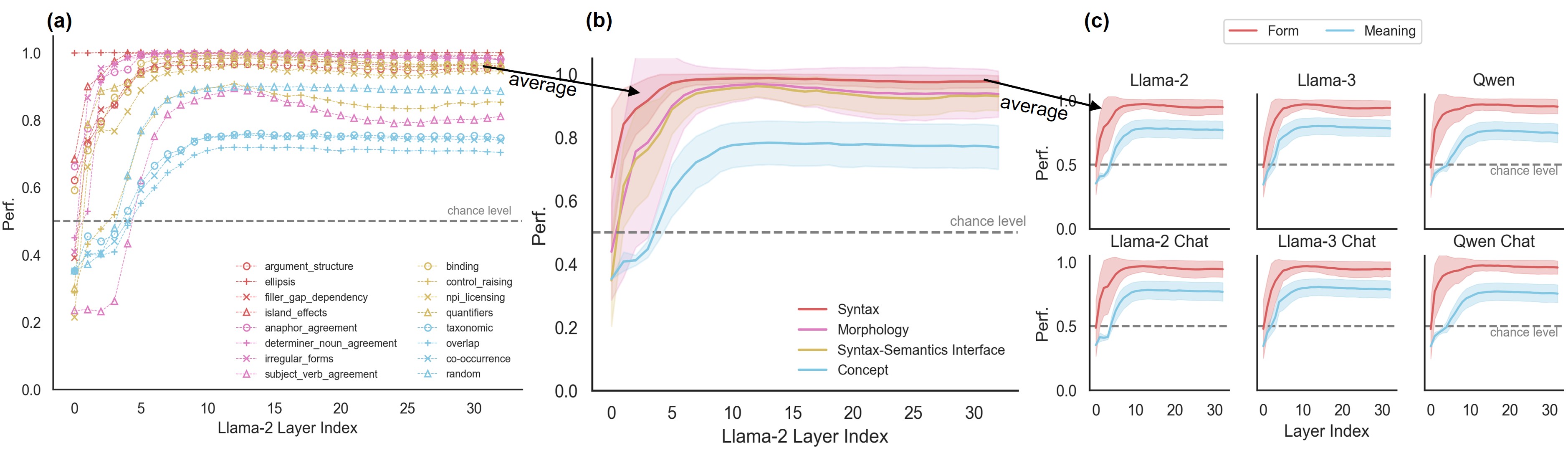}
    \caption{\textbf{(a)} Neurolinguistic probing performance for 16 tasks in Llama-2, including 4 syntax tasks, 4 morphology tasks, 4 syntax-semantics interface tasks, and 4 conceptual tasks.
\textbf{(b)} Average probing performance across the four linguistic categories in Llama-2. 
\textbf{(c)} Mean performance comparison between form-related tasks (syntax, morphology, syntax-semantics interface) and meaning-related tasks (concept), aggregated across all six models. 
 }
    \label{fig:4_level}
\end{figure*}

\begin{figure}[h]
    \centering
    \includegraphics[width=0.95\linewidth]{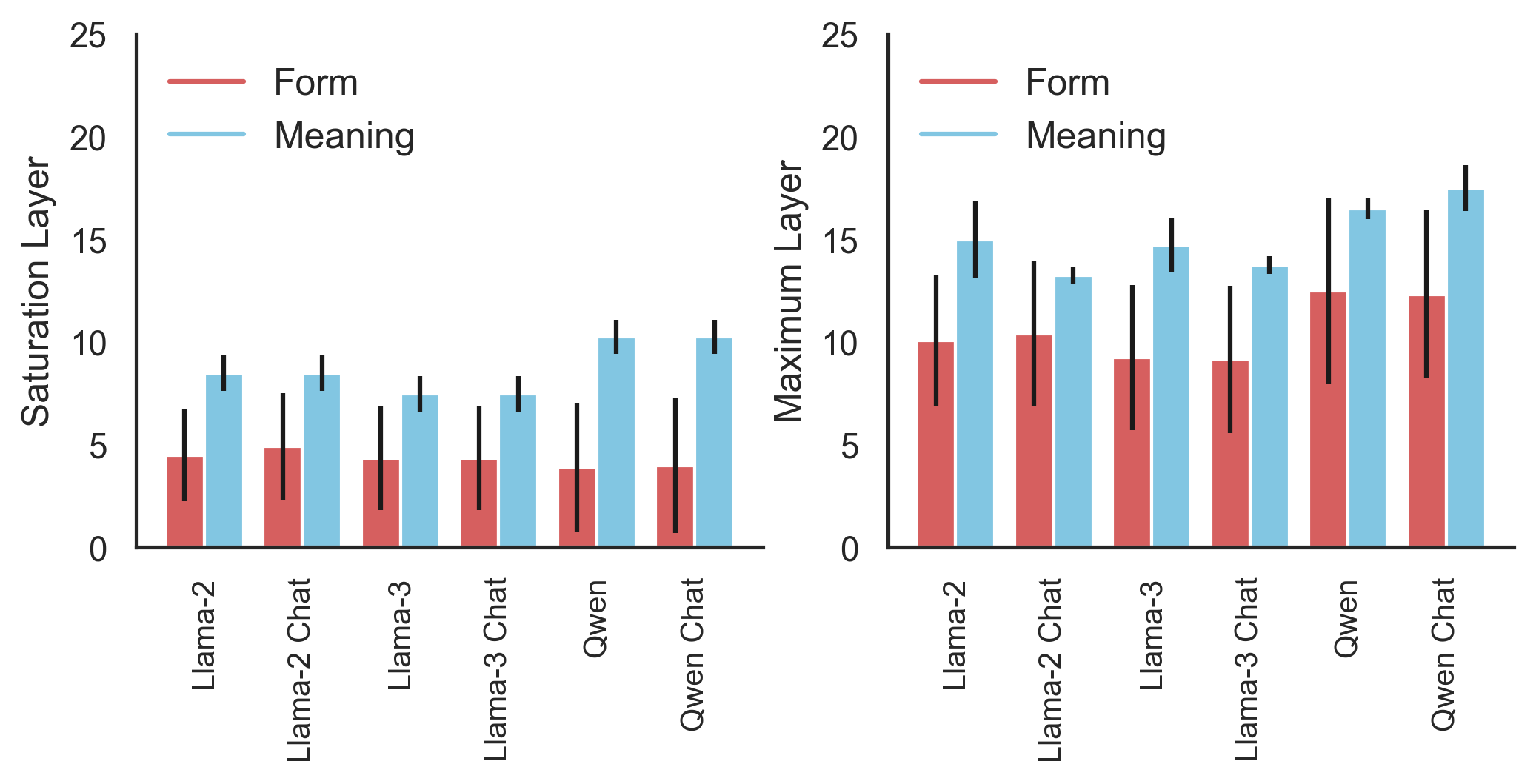}
    \caption{
Feature learning saturation layer (defined as the first layer reaching 95\% of peak performance) and the layer of maximum performance.
 }
    \label{fig:4_level_2}
\end{figure}

\begin{figure}[h]
    \centering
    \includegraphics[width=0.95\linewidth]{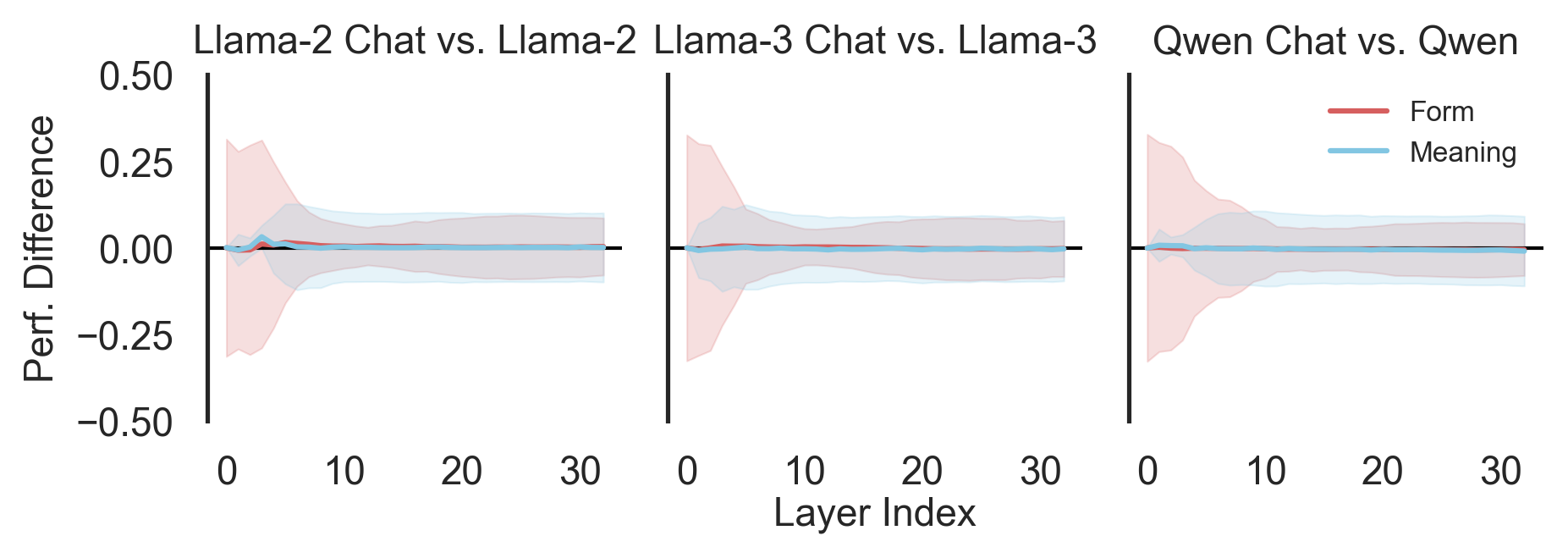}
    \caption{
Difference in probing performance between base and instruction-tuned models across all layers.
 }
    \label{fig:4_level_3}
\end{figure}

\subsection{Neurolinguistic Analysis\footnote{Raw results for English, Chinese, and German can be found in Figure~\ref{fig:decoding_english_all}, \ref{fig:decoding_chinese_all} and \ref{fig:decoding_german_all} in the Appendix.}}

\paragraph{Layer-wise unsupervised dynamics reveal gradual emergence of form features} Figure \ref{fig:tsne} illustrates the layer-wise differences between embeddings for grammatically correct and incorrect sentences. In early layers, the embedding difference appears scattered and unstructured, but as depth increases, they form clearer clusters, indicating a progressively refined sensitivity to syntactic correctness. By Layer 16 and beyond, distinct clusters emerge corresponding to syntax, morphology, and syntax-semantics interface. 
The results demonstrate that LLMs encode grammaticality judgments dynamically across layers, progressively structuring linguistic representations. Moreover, the formation of distinct clusters for different linguistic phenomena in the unsupervised analysis provides supporting evidence for subsequent supervised classification.

\paragraph{Gradual decline in encoding performance from structure to meaning.} The results in Figure~\ref{fig:4_level}-(c) show that the performance scores for conceptual understanding are significantly lower than those for grammatical understanding. This pattern is consistent across all six models, suggesting a universal characteristic of LLMs. Moreover, as illustrated in Figure~\ref{fig:4_level}-(a),(b), the encoding performance progressively declines from more structural tasks to more semantic tasks, spanning syntax, morphology, the syntax-semantic interface, and finally conceptual understanding. This highlights that LLMs encode features less effectively as the tasks shift from structure-focused to meaning-focused.




\paragraph{LLMs encode form earlier than meaning.} We compute the feature learning saturation and maximum layers for all 12 grammatical tasks and 4 conceptual tasks, averaging them to represent form and meaning, respectively.  As shown in Figure~\ref{fig:4_level_2}, the saturation and maximum layers for meaning are generally higher than those for form across all six models. This suggests that LLMs stabilize their encoding of grammatical features before conceptual features. 

\paragraph{Instruction tuning has minimal impact on the internal linguistic representations.} 
As Figure~\ref{fig:4_level_3} shows, performance differences for form and meaning remain near zero across all layers, indicating that instruction tuning minimally impacts internal linguistic representations, consistent with our psycholinguistic vs. neurolinguistic analysis.

\subsection{Multilingual analysis}
\begin{figure}[h]
    \centering
    \includegraphics[width=0.9\linewidth]{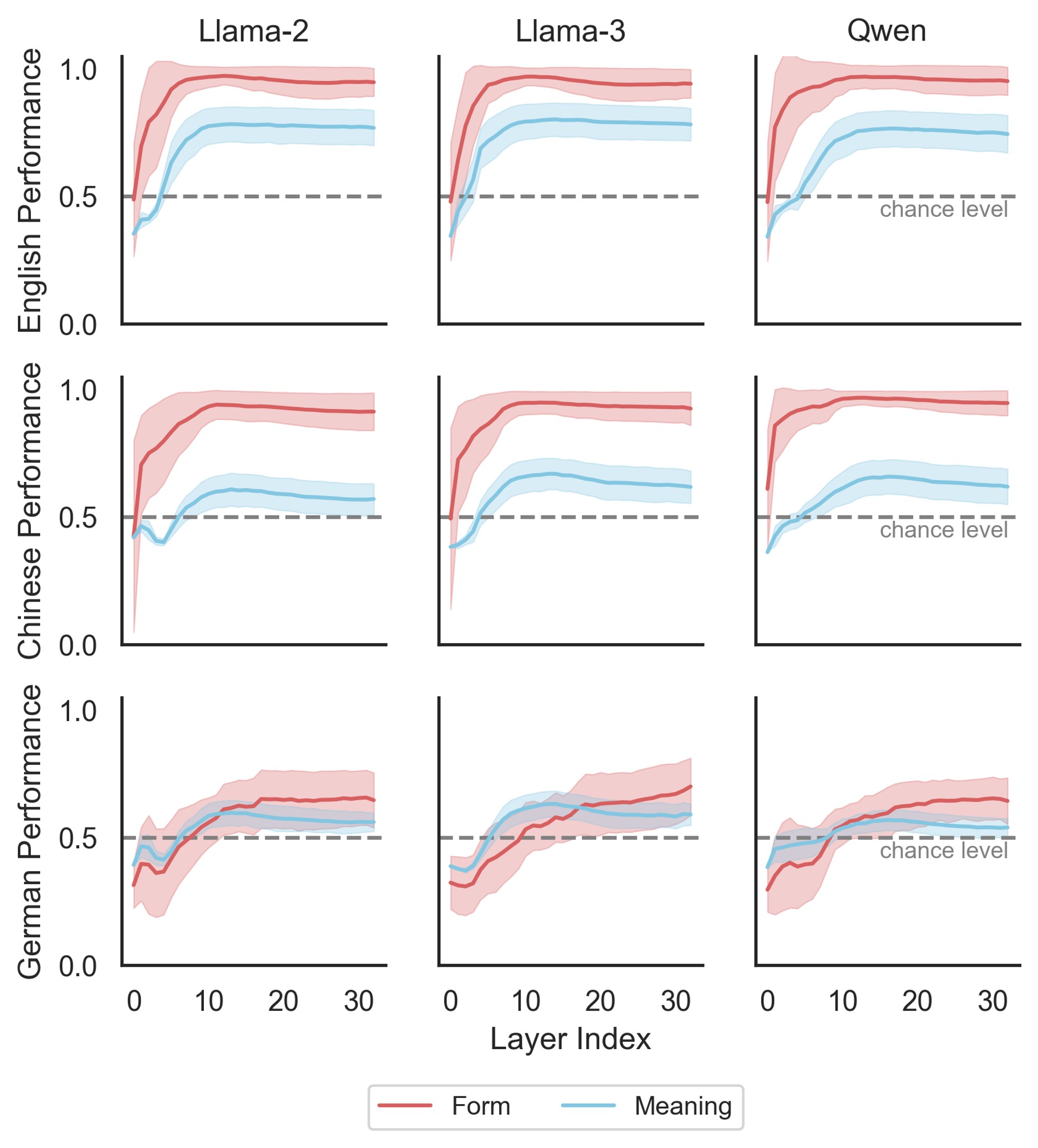}
    \caption{
    Neuro probing results for English, Chinese, and German.  
    }
    \label{fig:multiling}
\end{figure}
How does LLMs' understanding of meaning change when the form (language) varies? Since our COMPS-ZH and COMPS-DE datasets align with the concepts in the English COMPS dataset, we can explore whether LLMs' grasp of different linguistic forms for the same concept correlates with their understanding of meaning across languages. Our previous results suggest that instruction tuning has little influence on the internal representations. Therefore, we focus on the base LLMs here. 

From Figure~\ref{fig:multiling}, for all models and languages, form consistently achieves higher performance than meaning, indicating it's easier for LLMs to make a stronger grasp of structural elements compared to conceptual comprehension. Extended multilingual analysis can be found in Appendix \ref{sec:appendix_multilingual}.

\section{Discussion}



\paragraph{Language performance vs. competence: probing reveals deeper linguistic understanding than direct probability.}
Our results demonstrate that neurolinguistic probing uncovers linguistic competencies in LLMs that are not captured by psycholinguistic methods. While Meta performs the worst and Direct performs better, Neuro consistently outperforms both, revealing a systematic underestimation of competence when relying on output-based evaluations.

\citet{hu2023prompting} argued that Direct probability measurement, being more intrinsic than metalinguistic prompting, better reflects competence. However, our findings show that even Direct falls short of revealing the full extent of LLMs’ linguistic capabilities. Direct relies on the final output layer, which is highly optimized for next-word prediction and thus entangled with task-specific objectives. Prior studies \cite{hewitt2019structural,liu2019linguistic} have shown that syntactic and general linguistic information is often better represented in intermediate layers than in the final layer. \citet{waldis2024holmes} also emphasized that output correctness is an insufficient indicator of linguistic understanding, advocating for probing internal representations.

Our t-SNE visualizations corroborate this: clear linguistic clusters emerge in intermediate layers but dissolve in the final layer, reinforcing the view that the last layer is not optimal for assessing competence. These findings suggest that Direct, while more grounded than prompting, is still a limited proxy for internal knowledge.

In contrast, neurolinguistic probing inspects internal activation patterns across layers and tasks, uncovering the underlying representational structure of form and meaning, and further validates the discrepancy between performance and competence.

On the other hand, while Meta results underperform, this does not necessarily indicate that the LLMs lack the underlying linguistic competence. Instead, it may reflect limitations in information access, as suboptimal prompts can prevent models from exhibiting their full capabilities. Specifically, prompting failures do not always equate to a lack of encoded knowledge. This aligns with prior work \cite{firestone2020performance,lampinen2024can} emphasizing the need to distinguish performance conditions from underlying ability. 

Thus, we argue that probing, particularly when applied layer-wise, provides a more accurate and comprehensive assessment of linguistic competence than Direct probability alone.

\paragraph{Form and meaning: observations from Saussure's semiotics} 
Our results reveal that LLMs consistently learn linguistic form before they grasp meaning. This may suggest a developmental trajectory where statistical patterns in syntax and grammar are more readily captured by the model than conceptual understanding. Second, the models' formal competence is generally superior to their semantic competence. This is evident in their ability to decode grammaticality structures accurately but with less reliable conceptual accuracy.



We further observe a linear correlation between form and meaning competence, particularly when linguistic forms vary across languages while meaning remains constant. This suggests that LLMs' understanding of meaning might rely heavily on form, with conceptual representation anchored to formal structures rather than independent meaning comprehension. 
These results offer a semiotic and neurolinguistic explanation for LLMs' long-standing issue of generating ``confidently incorrect'' responses, i.e., hallucinations \cite{ji2023survey}.




\section{Conclusion}

This study adopts both psycho- and neuro-linguistic approaches to evaluating LLMs, revealing a distinction between linguistic performance and competence. 
Our results highlight the limitations of LLMs' semantic understanding and the need for future research to move beyond statistical correlations toward more grounded language representations. By introducing a cognitive neuroscience perspective, along with semiotics, we hope will inspire further research to deepen our understanding of the language capabilities of LLMs.

\section*{Limitations}
This study has several limitations that may impact the generalizability and comprehensiveness of our findings. First, we did not include experiments covering a wider range of languages, which restricts the cross-linguistic applicability of our results. Especially for the analysis and discussion in Section 5.3 on multilingual content, which highlights the necessity of constructing multilingual conceptual datasets.

Second, the evaluation results for German are notably poor, potentially due to the presence of very long sentences in the DistilLingEval dataset, which might have introduced challenges for the models. This underscores the need for constructing syntactic minimal pair datasets for German.

Additionally, our experiments were conducted using small-scale LLMs due to computational resource constraints. This may have introduced a bias in our findings, as larger-scale models could exhibit different behaviors. Future studies should explore larger models to validate and extend the generalizability of these results.

Lastly, the COMPS dataset used for assessing conceptual understanding is not sufficiently fine-grained, as it is limited to only four types of conceptual relationships. A more granular dataset could provide deeper insights into the nuances of how LLMs encode and process meaning. Future work should address this limitation by incorporating more diverse and detailed datasets.

\section*{Ethics Statement}
Our project has the potential to raise greater awareness within the computational linguistics community about the challenges faced by low-resource languages. By highlighting the unique linguistic features and limited computational tools available for these languages, we aim to inspire further research and the development of more inclusive language technologies that can better serve underrepresented linguistic communities.

\section*{Acknowledgements}
We thank the anonymous reviewers for their valuable advice and feedback.
This research was supported by DFG (German Research Foundation) grant SCHU 2246/14-1 and Munich Center for Machine Learning (MCML).

\bibliography{anthology,custom, references}

\begin{thebibliography}{59}
\expandafter\ifx\csname natexlab\endcsname\relax\def\natexlab#1{#1}\fi

\bibitem[{AI(2024)}]{meta2024llama3}
Meta AI. 2024.
\newblock Introducing meta llama 3: The most capable openly available llm to date.
\newblock \url{https://ai.meta.com/blog/meta-llama-3/}.

\bibitem[{Bai et~al.(2023)Bai, Bai, Chu, Cui, Dang, Deng, Fan, Ge, Han, Huang et~al.}]{bai2023qwen}
Jinze Bai, Shuai Bai, Yunfei Chu, Zeyu Cui, Kai Dang, Xiaodong Deng, Yang Fan, Wenbin Ge, Yu~Han, Fei Huang, et~al. 2023.
\newblock Qwen technical report.
\newblock \emph{arXiv preprint arXiv:2309.16609}.

\bibitem[{Belinkov(2022)}]{belinkov2022probing}
Yonatan Belinkov. 2022.
\newblock Probing classifiers: Promises, shortcomings, and advances.
\newblock \emph{Computational Linguistics}, 48(1):207--219.

\bibitem[{Belinkov and Glass(2019)}]{belinkov2019analysis}
Yonatan Belinkov and James Glass. 2019.
\newblock Analysis methods in neural language processing: A survey.
\newblock \emph{Transactions of the Association for Computational Linguistics}, 7:49--72.

\bibitem[{Bender and Koller(2020)}]{bender2020climbing}
Emily~M Bender and Alexander Koller. 2020.
\newblock Climbing towards nlu: On meaning, form, and understanding in the age of data.
\newblock In \emph{Proceedings of the 58th annual meeting of the association for computational linguistics}, pages 5185--5198.

\bibitem[{Beniwal et~al.(2024)Beniwal, D, and Singh}]{beniwal-etal-2024-cross}
Himanshu Beniwal, Kowsik D, and Mayank Singh. 2024.
\newblock \href {https://aclanthology.org/2024.findings-eacl.140} {Cross-lingual editing in multilingual language models}.
\newblock In \emph{Findings of the Association for Computational Linguistics: EACL 2024}, pages 2078--2128, St. Julian{'}s, Malta. Association for Computational Linguistics.

\bibitem[{Beyer et~al.(2021)Beyer, Lo{\'a}iciga, and Schlangen}]{beyer2021incoherence}
Anne Beyer, Sharid Lo{\'a}iciga, and David Schlangen. 2021.
\newblock Is incoherence surprising? targeted evaluation of coherence prediction from language models.
\newblock \emph{arXiv preprint arXiv:2105.03495}.

\bibitem[{Bloom(2002)}]{bloom2002children}
Paul Bloom. 2002.
\newblock \emph{How children learn the meanings of words}.
\newblock MIT press.

\bibitem[{Brennan(2022)}]{brennan2022language}
Jonathan~R Brennan. 2022.
\newblock \emph{Language and the brain: a slim guide to neurolinguistics}.
\newblock Oxford University Press.

\bibitem[{De~Saussure(1989)}]{de1989cours}
Ferdinand De~Saussure. 1989.
\newblock \emph{Cours de linguistique g{\'e}n{\'e}rale}, volume~1.
\newblock Otto Harrassowitz Verlag.

\bibitem[{Field(2004)}]{field2004psycholinguistics}
John Field. 2004.
\newblock \emph{Psycholinguistics: The key concepts}.
\newblock Psychology Press.

\bibitem[{Firestone(2020)}]{firestone2020performance}
Chaz Firestone. 2020.
\newblock Performance vs. competence in human--machine comparisons.
\newblock \emph{Proceedings of the National Academy of Sciences}, 117(43):26562--26571.

\bibitem[{Friederici(2011)}]{friederici2011brain}
Angela~D Friederici. 2011.
\newblock The brain basis of language processing: from structure to function.
\newblock \emph{Physiological reviews}, 91(4):1357--1392.

\bibitem[{Futrell et~al.(2019)Futrell, Wilcox, Morita, Qian, Ballesteros, and Levy}]{futrell2019neural}
Richard Futrell, Ethan Wilcox, Takashi Morita, Peng Qian, Miguel Ballesteros, and Roger Levy. 2019.
\newblock Neural language models as psycholinguistic subjects: Representations of syntactic state.
\newblock \emph{arXiv preprint arXiv:1903.03260}.

\bibitem[{Gauthier et~al.(2020)Gauthier, Hu, Wilcox, Qian, and Levy}]{gauthier2020syntaxgym}
Jon Gauthier, Jennifer Hu, Ethan Wilcox, Peng Qian, and Roger Levy. 2020.
\newblock Syntaxgym: An online platform for targeted evaluation of language models.
\newblock In \emph{Proceedings of the 58th Annual Meeting of the Association for Computational Linguistics: System Demonstrations}, pages 70--76.

\bibitem[{Gopnik and Wellman(2012)}]{gopnik2012reconstructing}
Alison Gopnik and Henry~M Wellman. 2012.
\newblock Reconstructing constructivism: causal models, bayesian learning mechanisms, and the theory theory.
\newblock \emph{Psychological bulletin}, 138(6):1085.

\bibitem[{Gupta et~al.(2015)Gupta, Boleda, Baroni, and Pad{\'o}}]{gupta2015distributional}
Abhijeet Gupta, Gemma Boleda, Marco Baroni, and Sebastian Pad{\'o}. 2015.
\newblock Distributional vectors encode referential attributes.
\newblock In \emph{Proceedings of the 2015 Conference on Empirical Methods in Natural Language Processing}, pages 12--21.

\bibitem[{Harnad(1990)}]{harnad1990symbol}
Stevan Harnad. 1990.
\newblock The symbol grounding problem.
\newblock \emph{Physica D: Nonlinear Phenomena}, 42(1-3):335--346.

\bibitem[{Haynes and Rees(2006)}]{haynes2006decoding}
John-Dylan Haynes and Geraint Rees. 2006.
\newblock Decoding mental states from brain activity in humans.
\newblock \emph{Nature reviews neuroscience}, 7(7):523--534.

\bibitem[{He et~al.(2024)He, Chen, Nie, Li, and Brennan}]{he2024decoding}
Linyang He, Peili Chen, Ercong Nie, Yuanning Li, and Jonathan~R Brennan. 2024.
\newblock Decoding probing: Revealing internal linguistic structures in neural language models using minimal pairs.
\newblock In \emph{Proceedings of the 2024 Joint International Conference on Computational Linguistics, Language Resources and Evaluation (LREC-COLING 2024)}, pages 4488--4497.

\bibitem[{Hewitt and Liang(2019)}]{hewitt2019designing}
John Hewitt and Percy Liang. 2019.
\newblock Designing and interpreting probes with control tasks.
\newblock \emph{arXiv preprint arXiv:1909.03368}.

\bibitem[{Hewitt and Manning(2019)}]{hewitt2019structural}
John Hewitt and Christopher~D Manning. 2019.
\newblock A structural probe for finding syntax in word representations.
\newblock In \emph{Proceedings of the 2019 Conference of the North American Chapter of the Association for Computational Linguistics: Human Language Technologies, Volume 1 (Long and Short Papers)}, pages 4129--4138.

\bibitem[{Hofstadter(1995)}]{hofstadter1995fluid}
Douglas~R Hofstadter. 1995.
\newblock \emph{Fluid concepts and creative analogies: Computer models of the fundamental mechanisms of thought.}
\newblock Basic books.

\bibitem[{Hu et~al.(2020)Hu, Gauthier, Qian, Wilcox, and Levy}]{hu2020systematic}
Jennifer Hu, Jon Gauthier, Peng Qian, Ethan Wilcox, and Roger~P Levy. 2020.
\newblock A systematic assessment of syntactic generalization in neural language models.
\newblock \emph{arXiv preprint arXiv:2005.03692}.

\bibitem[{Hu and Levy(2023)}]{hu2023prompting}
Jennifer Hu and Roger Levy. 2023.
\newblock Prompting is not a substitute for probability measurements in large language models.
\newblock In \emph{Proceedings of the 2023 Conference on Empirical Methods in Natural Language Processing}, pages 5040--5060.

\bibitem[{Ji et~al.(2023)Ji, Lee, Frieske, Yu, Su, Xu, Ishii, Bang, Madotto, and Fung}]{ji2023survey}
Ziwei Ji, Nayeon Lee, Rita Frieske, Tiezheng Yu, Dan Su, Yan Xu, Etsuko Ishii, Ye~Jin Bang, Andrea Madotto, and Pascale Fung. 2023.
\newblock Survey of hallucination in natural language generation.
\newblock \emph{ACM Computing Surveys}, 55(12):1--38.

\bibitem[{Kauf et~al.(2023)Kauf, Ivanova, Rambelli, Chersoni, She, Chowdhury, Fedorenko, and Lenci}]{kauf2023event}
Carina Kauf, Anna~A Ivanova, Giulia Rambelli, Emmanuele Chersoni, Jingyuan~Selena She, Zawad Chowdhury, Evelina Fedorenko, and Alessandro Lenci. 2023.
\newblock Event knowledge in large language models: the gap between the impossible and the unlikely.
\newblock \emph{Cognitive Science}, 47(11):e13386.

\bibitem[{Kemmerer(2022)}]{kemmerer2022cognitive}
David Kemmerer. 2022.
\newblock \emph{Cognitive neuroscience of language}.
\newblock Routledge.

\bibitem[{K{\"o}hn(2015)}]{kohn2015s}
Arne K{\"o}hn. 2015.
\newblock What’s in an embedding? analyzing word embeddings through multilingual evaluation.

\bibitem[{{Kriegeskorte} et~al.(2006){Kriegeskorte}, {Goebel}, and {Bandettini}}]{kriegeskorte2006information}
Nikolaus {Kriegeskorte}, Rainer {Goebel}, and Peter {Bandettini}. 2006.
\newblock Information-based functional brain mapping.
\newblock \emph{Proceedings of the National Academy of Sciences of the United States of America}, 103(10):3863--3868.

\bibitem[{Lai et~al.(2023)Lai, Ngo, Pouran Ben~Veyseh, Man, Dernoncourt, Bui, and Nguyen}]{lai-etal-2023-chatgpt}
Viet Lai, Nghia Ngo, Amir Pouran Ben~Veyseh, Hieu Man, Franck Dernoncourt, Trung Bui, and Thien Nguyen. 2023.
\newblock \href {https://doi.org/10.18653/v1/2023.findings-emnlp.878} {{C}hat{GPT} beyond {E}nglish: Towards a comprehensive evaluation of large language models in multilingual learning}.
\newblock In \emph{Findings of the Association for Computational Linguistics: EMNLP 2023}, pages 13171--13189, Singapore. Association for Computational Linguistics.

\bibitem[{Lake et~al.(2017)Lake, Ullman, Tenenbaum, and Gershman}]{lake2017building}
Brenden~M Lake, Tomer~D Ullman, Joshua~B Tenenbaum, and Samuel~J Gershman. 2017.
\newblock Building machines that learn and think like people.
\newblock \emph{Behavioral and brain sciences}, 40:e253.

\bibitem[{Lampinen(2024)}]{lampinen2024can}
Andrew Lampinen. 2024.
\newblock Can language models handle recursively nested grammatical structures? a case study on comparing models and humans.
\newblock \emph{Computational Linguistics}, pages 1--36.

\bibitem[{Linzen et~al.(2016)Linzen, Dupoux, and Goldberg}]{linzen2016assessing}
Tal Linzen, Emmanuel Dupoux, and Yoav Goldberg. 2016.
\newblock Assessing the ability of lstms to learn syntax-sensitive dependencies.
\newblock \emph{Transactions of the Association for Computational Linguistics}, 4:521--535.

\bibitem[{Liu et~al.(2019)Liu, Gardner, Belinkov, Peters, and Smith}]{liu2019linguistic}
Nelson~F Liu, Matt Gardner, Yonatan Belinkov, Matthew~E Peters, and Noah~A Smith. 2019.
\newblock Linguistic knowledge and transferability of contextual representations.
\newblock \emph{arXiv preprint arXiv:1903.08855}.

\bibitem[{Manning et~al.(2020)Manning, Clark, Hewitt, Khandelwal, and Levy}]{manning2020emergent}
Christopher~D Manning, Kevin Clark, John Hewitt, Urvashi Khandelwal, and Omer Levy. 2020.
\newblock Emergent linguistic structure in artificial neural networks trained by self-supervision.
\newblock \emph{Proceedings of the National Academy of Sciences}, 117(48):30046--30054.

\bibitem[{Marvin and Linzen(2018)}]{marvin-linzen-2018-targeted}
Rebecca Marvin and Tal Linzen. 2018.
\newblock \href {https://doi.org/10.18653/v1/D18-1151} {Targeted syntactic evaluation of language models}.
\newblock In \emph{Proceedings of the 2018 Conference on Empirical Methods in Natural Language Processing}, pages 1192--1202, Brussels, Belgium. Association for Computational Linguistics.

\bibitem[{Meng et~al.(2024)Meng, Liu, Joty, Xiong, Zhou, and Yavuz}]{SFRAIResearch2024}
Rui Meng, Ye~Liu, Shafiq~Rayhan Joty, Caiming Xiong, Yingbo Zhou, and Semih Yavuz. 2024.
\newblock \href {https://blog.salesforceairesearch.com/sfr-embedded-mistral/} {Sfr-embedding-mistral:enhance text retrieval with transfer learning}.
\newblock Salesforce AI Research Blog.

\bibitem[{Mesgarani and Chang(2012)}]{mesgarani2012selective}
Nima Mesgarani and Edward~F Chang. 2012.
\newblock Selective cortical representation of attended speaker in multi-talker speech perception.
\newblock \emph{Nature}, 485(7397):233--236.

\bibitem[{Misra et~al.(2023)Misra, Rayz, and Ettinger}]{misra2023comps}
Kanishka Misra, Julia Rayz, and Allyson Ettinger. 2023.
\newblock Comps: Conceptual minimal pair sentences for testing robust property knowledge and its inheritance in pre-trained language models.
\newblock In \emph{Proceedings of the 17th Conference of the European Chapter of the Association for Computational Linguistics}, pages 2920--2941.

\bibitem[{Mitchell and Krakauer(2023)}]{mitchell2023debate}
Melanie Mitchell and David~C Krakauer. 2023.
\newblock The debate over understanding in ai’s large language models.
\newblock \emph{Proceedings of the National Academy of Sciences}, 120(13):e2215907120.

\bibitem[{Nie et~al.(2023)Nie, Liang, Schmid, and Sch{\"u}tze}]{nie2023cross}
Ercong Nie, Sheng Liang, Helmut Schmid, and Hinrich Sch{\"u}tze. 2023.
\newblock Cross-lingual retrieval augmented prompt for low-resource languages.
\newblock In \emph{Findings of the Association for Computational Linguistics: ACL 2023}, pages 8320--8340.

\bibitem[{Nie et~al.(2024)Nie, Yuan, Ma, Schmid, F{\"a}rber, Kreuter, and Sch{\"u}tze}]{nie2024decomposed}
Ercong Nie, Shuzhou Yuan, Bolei Ma, Helmut Schmid, Michael F{\"a}rber, Frauke Kreuter, and Hinrich Sch{\"u}tze. 2024.
\newblock Decomposed prompting: Unveiling multilingual linguistic structure knowledge in english-centric large language models.
\newblock \emph{arXiv preprint arXiv:2402.18397}.

\bibitem[{Norman et~al.(2006)Norman, Polyn, Detre, and Haxby}]{norman2006beyond}
Kenneth~A Norman, Sean~M Polyn, Greg~J Detre, and James~V Haxby. 2006.
\newblock Beyond mind-reading: multi-voxel pattern analysis of fmri data.
\newblock \emph{Trends in cognitive sciences}, 10(9):424--430.

\bibitem[{Pinker(2009)}]{pinker2009language}
Steven Pinker. 2009.
\newblock \emph{Language learnability and language development: with new commentary by the author}, volume~7.
\newblock Harvard University Press.

\bibitem[{Sejnowski(2023)}]{sejnowski2023large}
Terrence~J Sejnowski. 2023.
\newblock Large language models and the reverse turing test.
\newblock \emph{Neural computation}, 35(3):309--342.

\bibitem[{Shi et~al.(2016)Shi, Padhi, and Knight}]{shi2016does}
Xing Shi, Inkit Padhi, and Kevin Knight. 2016.
\newblock Does string-based neural mt learn source syntax?
\newblock In \emph{Proceedings of the 2016 conference on empirical methods in natural language processing}, pages 1526--1534.

\bibitem[{Stouffer et~al.(1949)Stouffer, Suchman, DeVinney, Star, and Williams~Jr}]{stouffer1949american}
Samuel~A Stouffer, Edward~A Suchman, Leland~C DeVinney, Shirley~A Star, and Robin~M Williams~Jr. 1949.
\newblock \emph{The american soldier: Adjustment during army life.(studies in social psychology in world war ii), vol. 1}.
\newblock Princeton Univ. Press.

\bibitem[{Tenenbaum et~al.(2011)Tenenbaum, Kemp, Griffiths, and Goodman}]{tenenbaum2011grow}
Joshua~B Tenenbaum, Charles Kemp, Thomas~L Griffiths, and Noah~D Goodman. 2011.
\newblock How to grow a mind: Statistics, structure, and abstraction.
\newblock \emph{science}, 331(6022):1279--1285.

\bibitem[{Tenney et~al.(2019)Tenney, Xia, Chen, Wang, Poliak, McCoy, Kim, Van~Durme, Bowman, Das et~al.}]{tenney2019you}
Ian Tenney, Patrick Xia, Berlin Chen, Alex Wang, Adam Poliak, R~Thomas McCoy, Najoung Kim, Benjamin Van~Durme, Samuel~R Bowman, Dipanjan Das, et~al. 2019.
\newblock What do you learn from context? probing for sentence structure in contextualized word representations.
\newblock \emph{arXiv preprint arXiv:1905.06316}.

\bibitem[{Tomasello(2005)}]{tomasello2005constructing}
Michael Tomasello. 2005.
\newblock \emph{Constructing a language: A usage-based theory of language acquisition}.
\newblock Harvard university press.

\bibitem[{Touvron et~al.(2023{\natexlab{a}})Touvron, Lavril, Izacard, Martinet, Lachaux, Lacroix, Rozi{\`e}re, Goyal, Hambro, Azhar et~al.}]{touvron2023llama}
Hugo Touvron, Thibaut Lavril, Gautier Izacard, Xavier Martinet, Marie-Anne Lachaux, Timoth{\'e}e Lacroix, Baptiste Rozi{\`e}re, Naman Goyal, Eric Hambro, Faisal Azhar, et~al. 2023{\natexlab{a}}.
\newblock Llama: Open and efficient foundation language models.
\newblock \emph{arXiv preprint arXiv:2302.13971}.

\bibitem[{Touvron et~al.(2023{\natexlab{b}})Touvron, Martin, Stone, Albert, Almahairi, Babaei, Bashlykov, Batra, Bhargava, Bhosale et~al.}]{touvron2023llama2}
Hugo Touvron, Louis Martin, Kevin Stone, Peter Albert, Amjad Almahairi, Yasmine Babaei, Nikolay Bashlykov, Soumya Batra, Prajjwal Bhargava, Shruti Bhosale, et~al. 2023{\natexlab{b}}.
\newblock Llama 2: Open foundation and fine-tuned chat models.
\newblock \emph{arXiv preprint arXiv:2307.09288}.

\bibitem[{Traxler and Gernsbacher(2011)}]{traxler2011handbook}
Matthew Traxler and Morton~Ann Gernsbacher. 2011.
\newblock \emph{Handbook of psycholinguistics}.
\newblock Elsevier.

\bibitem[{Vamvas and Sennrich(2021)}]{vamvas2021limits}
Jannis Vamvas and Rico Sennrich. 2021.
\newblock On the limits of minimal pairs in contrastive evaluation.
\newblock \emph{arXiv preprint arXiv:2109.07465}.

\bibitem[{Waldis et~al.(2024)Waldis, Perlitz, Choshen, Hou, and Gurevych}]{waldis2024holmes}
Andreas Waldis, Yotam Perlitz, Leshem Choshen, Yufang Hou, and Iryna Gurevych. 2024.
\newblock Holmes a benchmark to assess the linguistic competence of language models.
\newblock \emph{Transactions of the Association for Computational Linguistics}, 12:1616--1647.

\bibitem[{Wang et~al.(2023)Wang, Haddow, and Birch}]{wang2023retrieval}
Weixuan Wang, Barry Haddow, and Alexandra Birch. 2023.
\newblock Retrieval-augmented multilingual knowledge editing.
\newblock \emph{arXiv preprint arXiv:2312.13040}.

\bibitem[{Warstadt et~al.(2020)Warstadt, Parrish, Liu, Mohananey, Peng, Wang, and Bowman}]{warstadt2020blimp}
Alex Warstadt, Alicia Parrish, Haokun Liu, Anhad Mohananey, Wei Peng, Sheng-Fu Wang, and Samuel~R Bowman. 2020.
\newblock Blimp: The benchmark of linguistic minimal pairs for english.
\newblock \emph{Transactions of the Association for Computational Linguistics}, 8:377--392.

\bibitem[{Xiang et~al.(2021)Xiang, Yang, Li, Warstadt, and Kann}]{xiang2021climp}
Beilei Xiang, Changbing Yang, Yu~Li, Alex Warstadt, and Katharina Kann. 2021.
\newblock Climp: A benchmark for chinese language model evaluation.
\newblock \emph{arXiv preprint arXiv:2101.11131}.

\end{thebibliography}
\bibliographystyle{acl_natbib}

\appendix
\section{Neuro Probing's Cognitive Science Background}
\label{sec:appendix_paradigm}
Psycholinguistics often involves examining real-time language processing, linguistic knowledge storage, and language acquisition, using behavioral experimental methods such as reading times and eye-tracking. Neurolinguistics, on the other hand, focuses on the neural basis of language, employing techniques such as functional magnetic resonance imaging (fMRI), electroencephalography (EEG), and magnetoencephalography (MEG) to map linguistic functions to specific brain regions and to investigate how neural activity correlates with linguistic tasks.

While psycholinguistics aims to reveal \textit{mental} processes underlying language use, neurolinguistics seeks to uncover the \textit{neural} pathways that implement these processes. 

Our minimal pair probing is inspired by cognitive neuroscience. In the field of neurolinguistics, decoding analysis has become a fundamental technique in cognitive neuroscience. It tries to extract information encoded in neural patterns \citep{kriegeskorte2006information}. It trains a classifier to predict the properties of the stimulus (e.g.\ a particular image or word input), from the neural responses. If the accuracy of the trained classifier is significantly better than chance, we conclude that the neural data encodes information about the predicted stimulus properties
\citep{norman2006beyond,haynes2006decoding}.

\citet{mesgarani2012selective} is a representative work employing decoding analysis in the realm of language and speech. They use electrocorticography (ECoG) to record neural responses from subjects who are listening to speech. Leveraging decoding analysis, they were able to differentiate between neural patterns induced by attended speech and those elicited by background speech (to be ignored by the test persons), thereby highlighting the perceptual differences between the two speech stimulus conditions.


While psycholinguistic approaches provide valuable insights into LLMs' functional capabilities, they often fall short in revealing the underlying mechanisms of language processing. The opaque nature of neural network structures means that performance on external tasks does not necessarily reflect the internal cognitive processes at play. This gap necessitates a neurolinguistic approach to gain a deeper understanding of how LLMs encode and process language. 
\section{Dataset Details}
\label{sec:appendix_dataset}

For each language, we use one dataset for grammatical minimal pairs and one dataset for conceptual minimal pairs. 
\subsection{Form: BLiMP, CLiMP, and DistilLingEval}
\paragraph{BLiMP} BLiMP~\citep{warstadt2020blimp} is a comprehensive English dataset of grammatical minimal pairs. It consists of minimal pairs for 13 higher-level linguistic phenomena in the English language, further divided into 67 distinct realizations, called paradigms. Each paradigm comprises 1,000 individual minimal pairs, resulting in a total corpus size of 67,000 data points.

\paragraph{CLiMP} 
CLiMP~\citep{xiang2021climp} is a corpus of Chinese grammatical minimal pairs consisting of 16 datasets, each containing 1,000 sentence pairs. CLiMP covers 9 major Chinese language phenomena in total, fewer than the BLiMP dataset due to the less inflectional nature of Mandarin Chinese. The vocabulary of the CLiMP dataset is based on the translation of the BLiMP dataset, with words and features specific to Chinese added.

\paragraph{DistilLingEval} DistilLingEval~\citep{vamvas2021limits} is a dataset of German grammatical minimal pairs. It consists of minimal pairs for eight German linguistic phenomena. This dataset contains 82,711 data samples in total.

\subsection{Meaning: COMPS, COMPS-ZH, and COMPS-DE}
\paragraph{COMPS} COMPS~\citep{misra2023comps} is an English dataset of conceptual minimal pairs for testing an LLM's knowledge of everyday concepts (e.g., \emph{a \textbf{beaver}/*gorilla has a flat tail}). This dataset contains 49,340 sentence pairs, constructed using 521 concepts and 3,592 properties. Concepts in the pairs constitute 4 types of knowledge relationships: taxonomy, property norms, co-occurrence, and random.



\paragraph{COMPS-ZH and COMPS-DE}
COMPS-DE and COMPS-ZH are newly developed datasets featuring conceptual minimal pairs in Chinese and German, derived from the English COMPS dataset~\citep{misra2023comps}. 
In the realm of multilingual NLP research, it is a common practice to extend English datasets to other languages using human translation, machine translation, or translation assisted by LLMs~\citep{nie2023cross,wang2023retrieval,beniwal-etal-2024-cross}.

In this study, to create COMPS-DE and COMPS-ZH from the original English COMPS, we employed a hybrid approach that integrated process machine translation with meticulous human verification. 

Specifically, we translated the concepts and properties of the English COMPS individually, subsequently merging them to form complete sentences and compose conceptual minimal pairs. The translation process began with the use of the Google Translate API\footnote{https://cloud.google.com/translate}, which provided initial translations of concepts and properties into German and Chinese. 

Following this, native speakers of Chinese and German manually checked and refined these translations to ensure accuracy and quality. The manual review emphasized two main areas: accuracy of concepts and grammatical consistency of properties. For concepts, the focus was on correcting ambiguities that might arise from machine translation. For properties, attention was given to maintaining grammatical consistency with the original English text, such as ensuring subject-verb agreement, which is particularly challenging in German translations.

In summary, out of 521 concepts, manual corrections were made to 57 entries in the Chinese dataset and 49 in the German dataset. Similarly, out of 3,592 properties, 713 required manual corrections in the Chinese dataset, and 512 in the German dataset. This rigorous process was essential for preserving the integrity and reliability of the translated datasets.

\section{Model Details}
\label{sec:appendix_models}
In our experiments, we used three open-source LLMs, two English-centric LLMs (Llama2 and Llama3), and one multilingual LLM (Qwen) with a focus on English and Chinese.
These models were trained on different amounts of English, Chinese, and German data (see Table \ref{tab:models}). 

\begin{table}[h]
    \centering
    \resizebox{1\linewidth}{!}{
    \begin{tabular}{llll}
    \hline
    \textbf{Resource Level} & \textbf{Llama2} & \textbf{Llama3} & \textbf{Qwen} \\ \hline
    English                 & High              & High              & High            \\
    Chinese                 & Mid              & Mid              & High           \\
    German                  & Low             & Low             & Low           \\ \hline
    \end{tabular}
    }
    \caption{Resource level for different languages across three LLMs. Note the resource levels are qualitative assessments based on available information, as specific quantitative data is not provided by the developers.}
    \label{tab:models}
\end{table}

\paragraph{Llama2 and Llama3} Llama2~\citep{touvron2023llama2} and Llama3~\citep{meta2024llama3} are two English-centric LLMs which represent an advanced iteration of the Llama foundation models developed by Meta AI~\citep{touvron2023llama}. The Llama
madels were 
trained on publicly available corpora predominantly in English. 
Despite this focus, Llama models are also exposed to a limited amount of multilingual data. Llama 1, for example, is pretrained on an extensive scale of corpora comprising over 1.4 trillion tokens, of which less than 4.5\% constitute multilingual data from 20 different languages. Llama 2 expands this linguistic diversity, featuring 27 languages, each representing more than 0.005\% of the pertaining data. Therefore, English-centric models harness multilingual abilities~\citep{lai-etal-2023-chatgpt}. In this work, we use Llama2-7B and Llama3-8B for our experiments.

\paragraph{QWen} QWen is a series of LLMs developed by Alibaba Inc.~\cite{bai2023qwen}. Qwen was trained on 2-3 trillion tokens of multilingual pre-training data. It is essentially a multilingual LLM with a focus on English and Chinese. We use the Qwen-7B model in our experiments.

\section{Supplemented Results for English}
Some noteworthy points from Figure~\ref{fig:4_level}-(a) include: \\

\noindent\textbf{1)}
For the ellipsis task, especially, local features in layer 0 (the embedding layer before the Transformer structure) already provide sufficient linguistic information to accomplish the task without any contextual information.\\
\noindent\textbf{2)} Among the conceptual tasks, the random relationship shows significantly higher accuracy compared to the other three conceptual relationships, suggesting that LLMs find it challenging to distinguish between similar concepts.

\subsubsection*{Compared to Llama2, Llama3 won't improve internal grammatical capabilities much, but will learn concepts better and faster.}
\begin{figure}[h]
\centering
\includegraphics[width=1\linewidth]{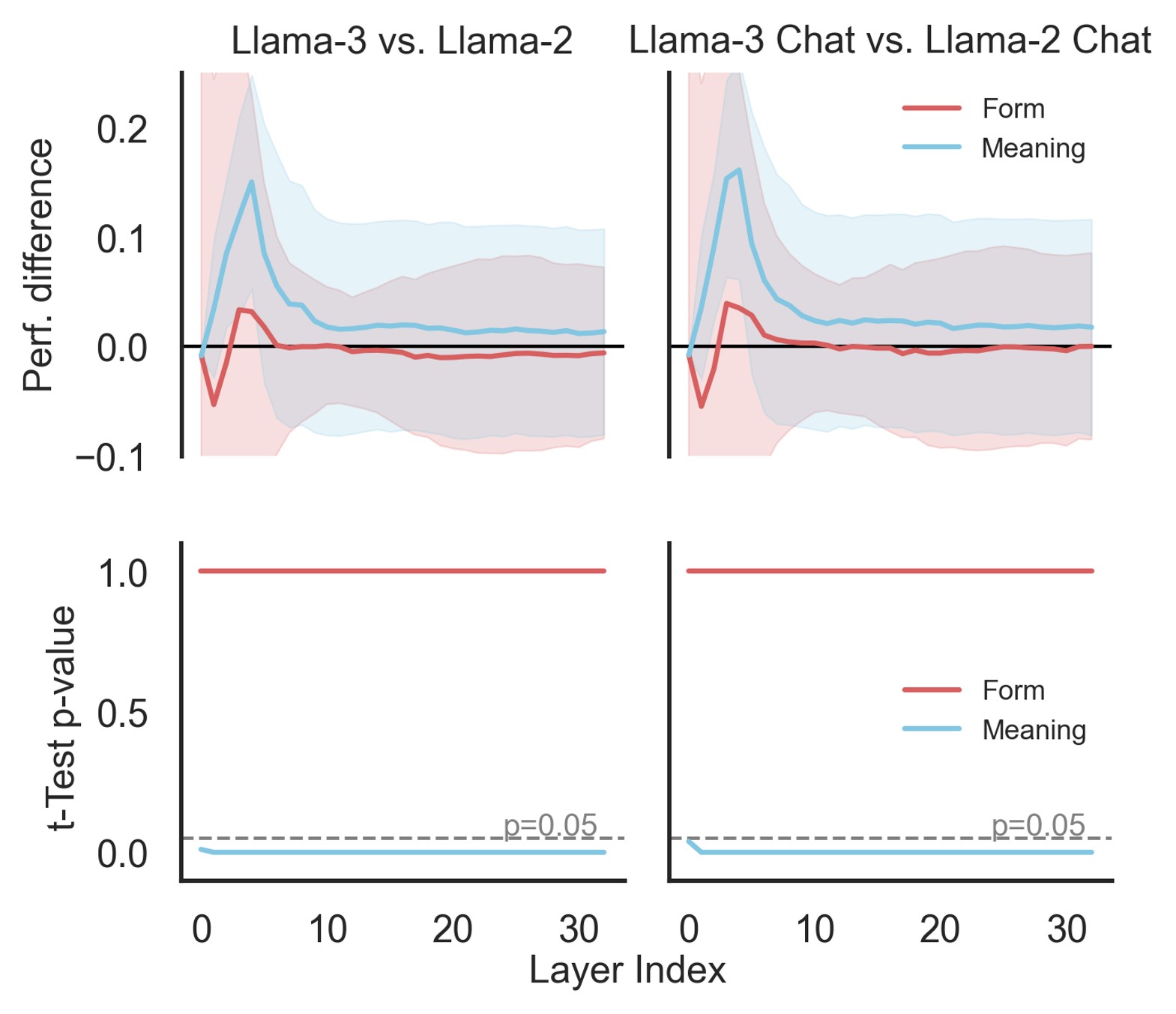}
\caption{Top: Performance difference between Llama3 and Llama2. Bottom: Layer-wise t-test results. T-tests were first performed separately on each linguistic task, and then Stouffer's Z-score method \cite{stouffer1949american} was employed to aggregate the final p-value at the condition level.}
\label{fig:difference_32}
\end{figure}


Figure~\ref{fig:difference_32} shows the layer-wise performance differences between Llama3 and Llama2, as well as between their chat versions. The red curves (meaning) exhibit a notable positive difference in the early layers, indicating that Llama3 has better conceptual learning capabilities compared to Llama2. The blue curves (form) remain close to zero across all layers, suggesting that there is no significant improvement in grammatical capabilities in Llama3 compared to Llama2. The t-test statistics in Figure~\ref{fig:difference_32} further support these results. 

\begin{figure}[h]
\centering
\includegraphics[width=0.8\linewidth]{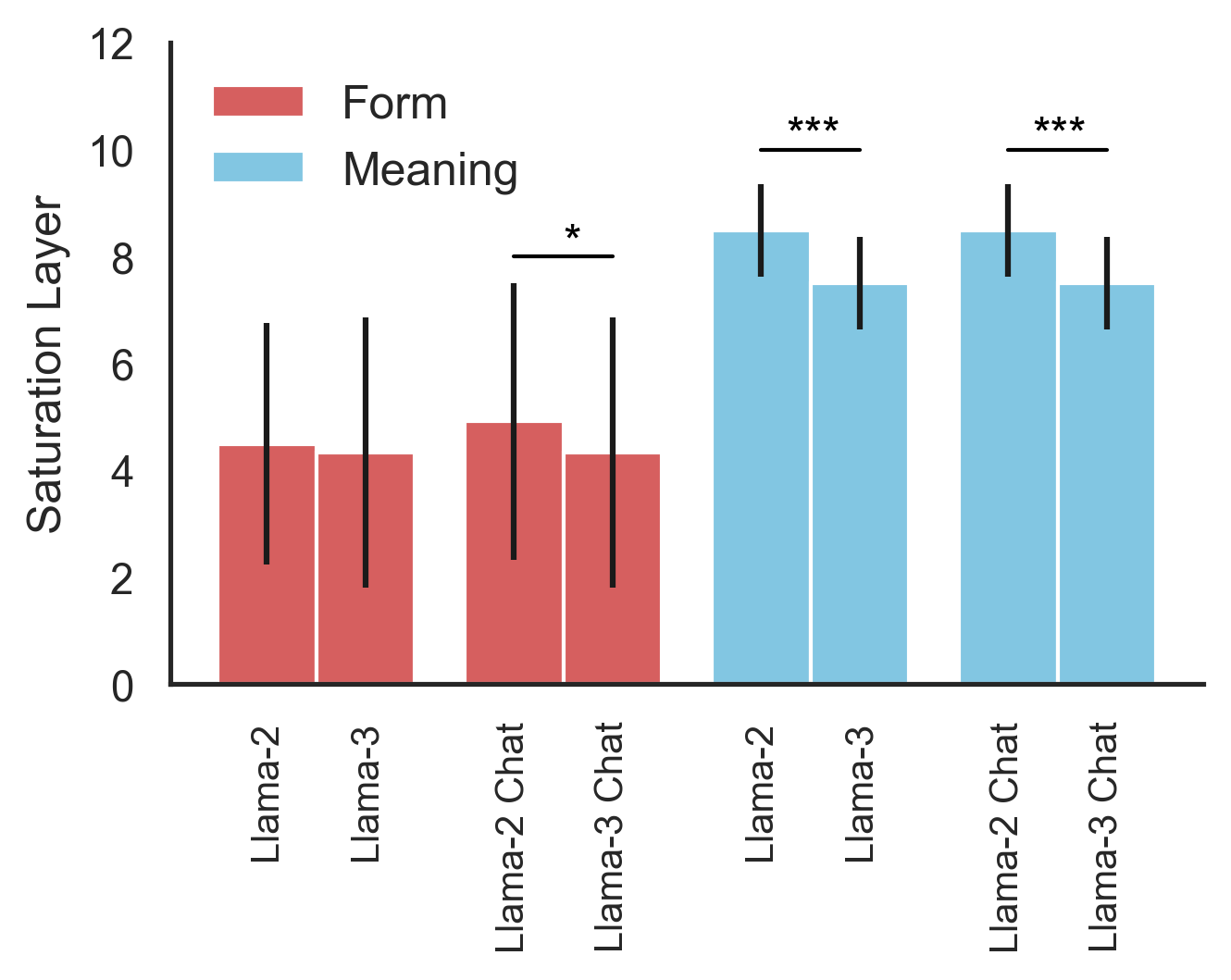}
\caption{T-test results between Llama2 and Llama3 feature learning saturation layer. The symbols `***', and `*' denote t-test p-values less than 0.001 and 0.05, respectively.  }
\label{fig:ttest_saturation}
\end{figure}

Figure~\ref{fig:ttest_saturation} compares the feature learning saturation layers between Llama2 and Llama3.  The results for form learning (blue bars) do not differ significantly between Llama2 and Llama3, and weakly significantly between Llama2\_chat and Llama3\_chat. However, the results for meaning learning (blue bars) are both highly significant, indicating that Llama3 requires fewer layers to encode conceptual features than Llama2. This suggests that Llama3
comprehends sentences faster.
\section{Supplemented Results for Multilingual Analysis}
\label{sec:appendix_multilingual}
\begin{figure}[h]
    \centering
    \includegraphics[width=0.7\linewidth]{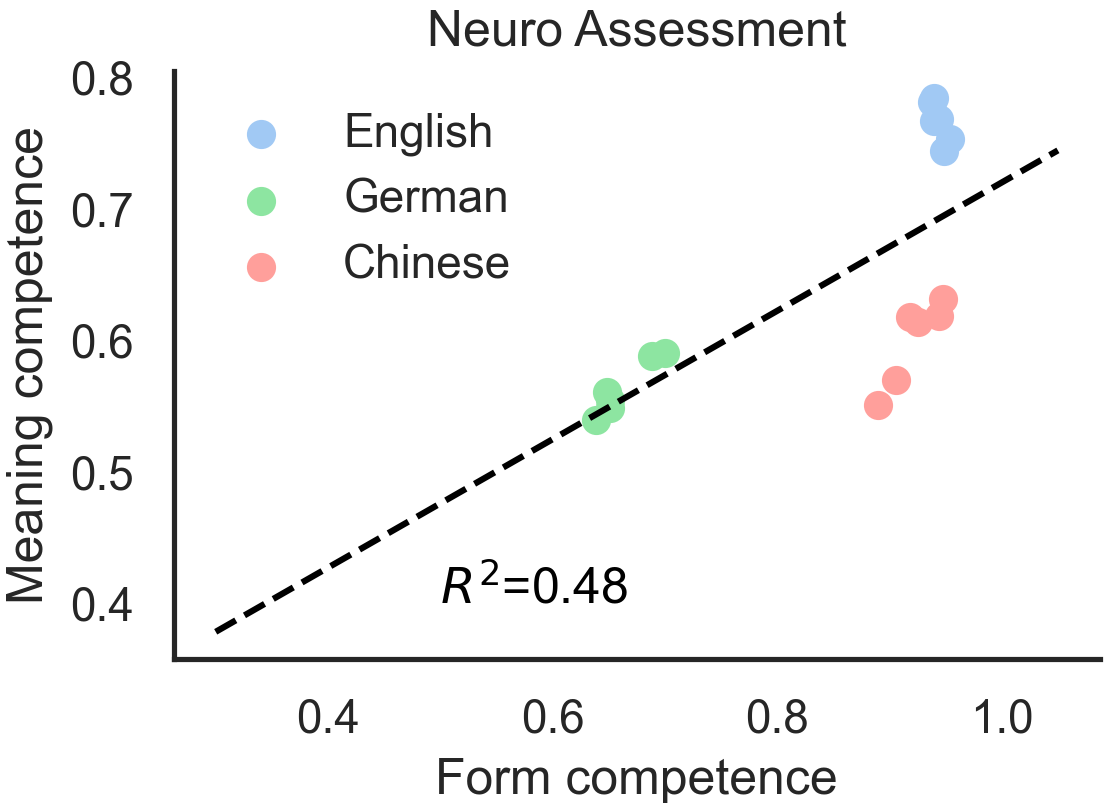}
    \caption{
    Correlation between meaning competence and form competence.}
    \label{fig:multiling_2}
\end{figure}

\paragraph{Meaning competence is correlated to form competence.} Figure \ref{fig:multiling_2} illustrates the relationship between Form competence (x-axis) and Meaning competence (y-axis) across English, German, and Chinese in the neuro assessment for LLMs. The positive correlation (R² = 0.48) suggests that higher meaning competence generally corresponds to higher form competence.

\paragraph{Llama's performance on Chinese.} 
Despite Chinese not being the primary training language of the Llama2 models, they still perform well in encoding Chinese form/grammar. However, both Qwen, which is primarily trained on Chinese, and the Llama models show relatively poor performance in understanding Chinese meaning/concepts.

\paragraph{Improvement in Llama3 for Chinese Semantics.} Llama3 shows some improvement over Llama2 in understanding Chinese semantics, as indicated by the slightly higher red curve in the middle row. The improvement in syntactic understanding is minimal.

\paragraph{Qwen's Faster Syntax Learning but Slower Semantic Encoding for Chinese.} Compared to the Llama models, Qwen learns Chinese grammar faster, as indicated by the sharper rise of the blue curve. However, it encodes Chinese semantics more slowly, evidenced by the larger gap between the form and meaning curves in the early layers.

\paragraph{Poor Performance for German.}
For German, a low-resource language, all three models perform poorly.
Despite Chinese not being a primary training language for the Llama models, their performance is relatively decent, suggesting that the actual proportion of German training data might be much smaller. This highlights differences in the resource allocation for the three languages.
\paragraph{Form needs less data to capture compared to meaning.} From Table \ref{tab:models}, Chinese is classified as a mid-resourced language for Llama, yet it achieves high form competence (but low meaning competence), suggesting that capturing form requires less data than meaning.
\section{Supplemented Discussion}
\paragraph{Developmental difference between human and machine intelligence.} From a perspective of developmental psychology, human kids typically acquire conceptual understanding before mastering grammar \cite{bloom2002children, tomasello2005constructing}. \citet{pinker2009language}'s semantic bootstrapping hypothesis posits that children initially learn vocabulary through semantic information and then use this semantic knowledge to infer syntactic structures. In contrast, our results indicate that LLMs learn grammar before meaning. Human intelligence is a combination of statistical inference and causal reasoning, whereas LLMs' intelligence is more likely a result of statistical inference \cite{tenenbaum2011grow, gopnik2012reconstructing, lake2017building}. Given this nature, the fact that LLMs learn form first might be because grammatical patterns are easier to statistically capture compared to meaning. 
\paragraph{Symbol grounding problem and the quest for human-like intelligence} In human language, the relationship between the signifier and the signified is often flexible and context-dependent, allowing for a more independent connection between syntax and semantics\cite{harnad1990symbol}. Human cognitive development typically involves acquiring conceptual understanding first, followed by the learning of rules and syntax. In contrast, our study shows that LLMs grasp syntax before meaning, relying on statistical correlations within formal structures to infer semantic content. This difference highlights a fundamental divergence between human and machine intelligence, as LLMs do not possess an inherent understanding of meaning detached from the formal structures they analyze.

These observations suggest that, for LLMs to develop human-like intelligence, they must transcend mere statistical pattern recognition. This will likely require the integration of world knowledge and grounded experiences that go beyond linguistic inputs. To achieve a more robust form of artificial intelligence that mirrors human cognition, models must be able to ground symbols in real-world contexts, establishing a basis for genuine understanding \cite{tenenbaum2011grow, lake2017building} As it stands, the symbol grounding problem remains a significant barrier, and addressing it may be essential for constructing systems capable of true human-like reasoning and understanding.
\begin{figure*}[h]
    \centering
    \includegraphics[width=0.9\linewidth]{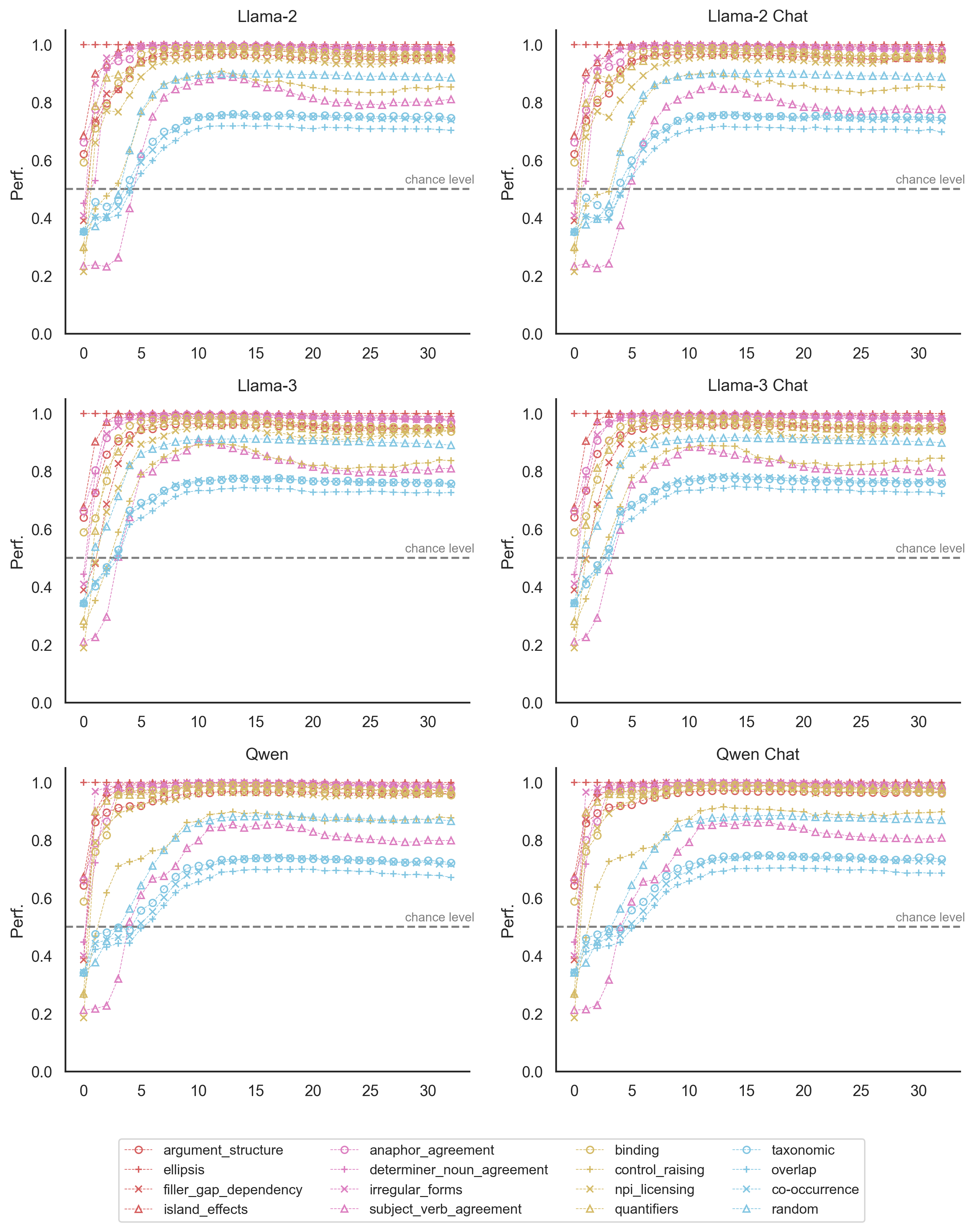}
    \caption{Detailed English decoding results on 6 models.}
    \label{fig:decoding_english_all}
\end{figure*}

\begin{figure*}[h]
    \centering
    \includegraphics[width=1\linewidth]{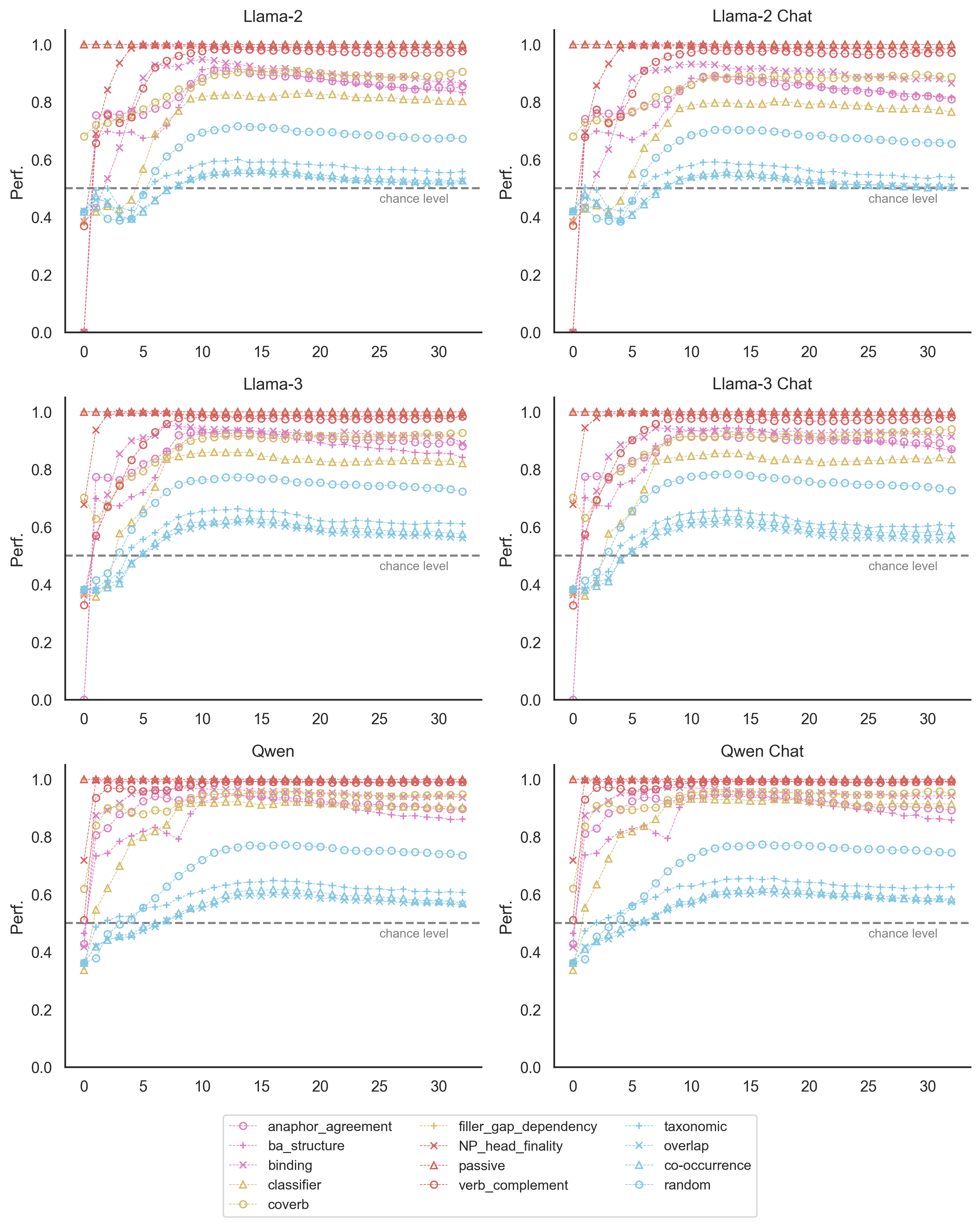}
    \caption{Detailed Chinese decoding results on 6 models. Notice that the pink, orange, and blue curves don't denote morphology or semantics as those in English do. They are made just to make it easier to distinguish in the figure. All non-red curves represent grammatical tasks and red curves represent conceptual tasks.}
    \label{fig:decoding_chinese_all}
\end{figure*}

\begin{figure*}[h]
    \centering
    \includegraphics[width=1\linewidth]{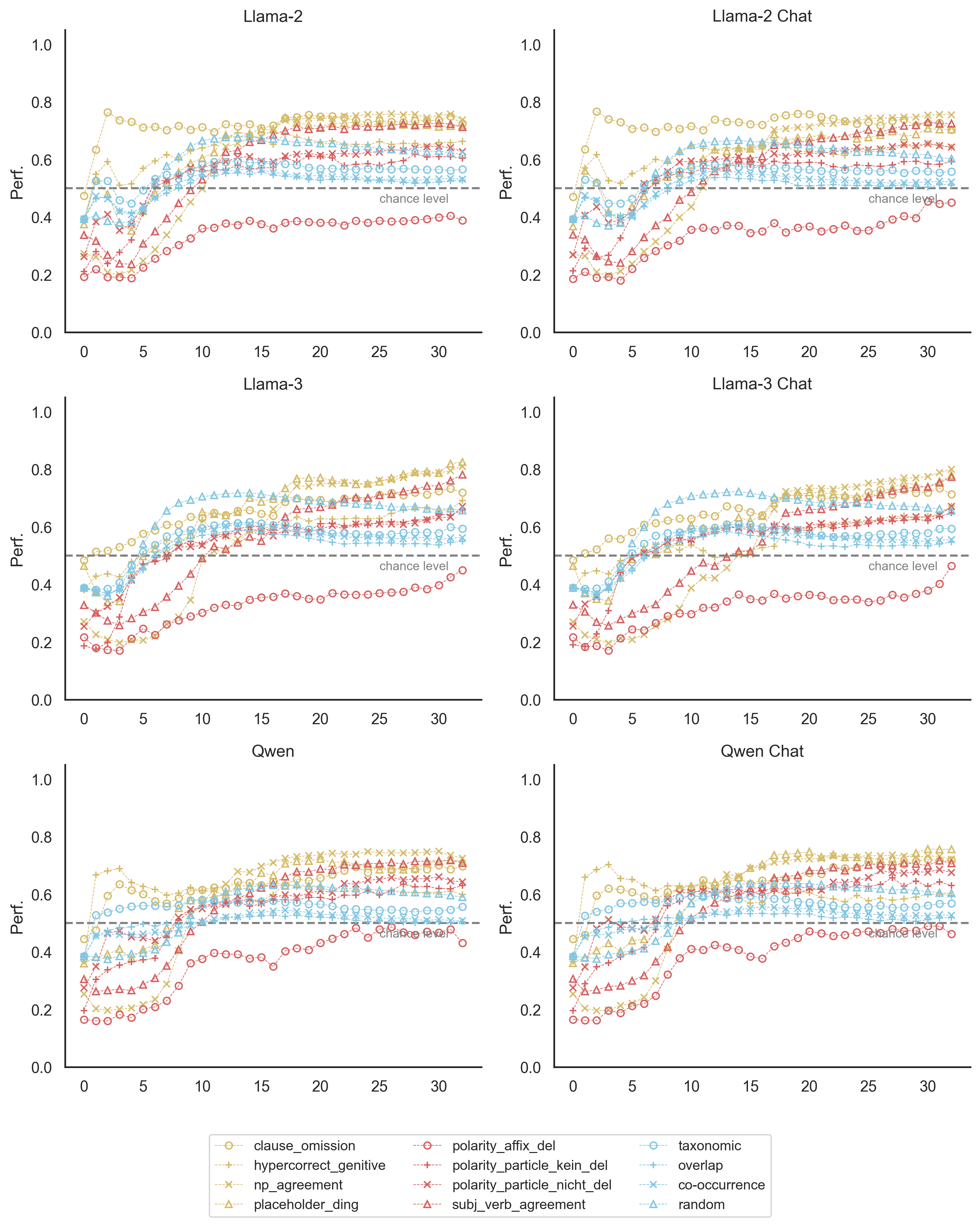}
    \caption{Detailed German decoding results on 6 models. All non-red curves are grammatical tasks, and red curves are conceptual tasks.}
    \label{fig:decoding_german_all}
\end{figure*}

\end{document}